\documentclass{article}

\usepackage{microtype}
\usepackage{graphicx}
\usepackage{subfigure}
\usepackage{booktabs} % for professional tables

\usepackage{hyperref}

\usepackage[accepted]{icml2025}

\usepackage{amsmath}
\usepackage{amssymb}
\usepackage{mathtools}
\usepackage{amsthm}

\usepackage{multirow}
\usepackage{bbm}

\usepackage[capitalize,noabbrev]{cleveref}

\definecolor{green(pigment)}{rgb}{0.1607, 0.3843, 0.0941}
\definecolor{blue(pigment)}{rgb}{0., 0.1484, 0.6992}
\definecolor{blue(back)}{rgb}{0.9058, 0.9019, 0.9725}
\definecolor{orange(pigment)}{rgb}{0.6, 0.298, 0.}
\theoremstyle{plain}
\newtheorem{theorem}{Theorem}[section]

\newtheorem{lemma}[theorem]{Lemma}

\newtheorem{mypro}[theorem]{Proof}
\theoremstyle{definition}

\theoremstyle{remark}

\definecolor{green(pigment)}{rgb}{0.1607, 0.3843, 0.0941}
\definecolor{blue(pigment)}{rgb}{0., 0.1484, 0.6992}
\hypersetup{hidelinks}
\hypersetup{colorlinks,linkcolor=blue(pigment),citecolor=green(pigment),urlcolor=black}

\usepackage{colortbl}

\newcommand{\grr}{\cellcolor[gray]{.95}}

\usepackage[textsize=tiny]{todonotes}

\usepackage{amsmath,amsfonts,bm}

\def\eqref#1{equation~\ref{#1}}
\def\1{\bm{1}}

\def\vv{{\mathbf{v}}}

\DeclareMathAlphabet{\mathsfit}{\encodingdefault}{\sfdefault}{m}{sl}
\SetMathAlphabet{\mathsfit}{bold}{\encodingdefault}{\sfdefault}{bx}{n}

\newcommand{\E}{\mathbb{E}}

\newcommand{\KL}{D_{\mathrm{KL}}}

\DeclareMathOperator*{\argmax}{arg\,max}

\newcommand{\x}{\mathbf{x}}
\newcommand{\y}{\mathbf{y}}
\newcommand{\w}{\mathbf{w}}
\newcommand{\W}{\mathbf{W}}

\newcommand{\X}{\mathcal{X}_B}
\newcommand{\Y}{\mathcal{Y}}
\newcommand{\St}{\mathcal{S}}
\newcommand{\D}{\mathcal{D}}
\newcommand{\G}{\mathcal{G}}
\newcommand{\C}{\mathcal{C}}
\newcommand{\Lc}{\mathcal{L}}
\newcommand{\act}{\phi}
\newcommand{\vecc}{\text{vec}}
\newcommand{\ul}{\mathbf{u}}

\newcommand{\pr}{\mathbb{P}}

\icmltitlerunning{Submission and Formatting Instructions for ICML 2025}

\begin{document}

\twocolumn[
\icmltitle{Principal Eigenvalue Regularization for Improved Worst-Class Certified Robustness of Smoothed Classifiers}




\begin{icmlauthorlist}
\icmlauthor{Gaojie Jin}{yyy}
\icmlauthor{Tianjin Huang}{yyy}
\icmlauthor{Ronghui Mu}{yyy}
\icmlauthor{Xiaowei Huang}{comp}
\end{icmlauthorlist}

\icmlaffiliation{yyy}{Department of Computer Science, University of Exeter, UK}
\icmlaffiliation{comp}{Department of Computer Science, University of Liverpool, UK}

\icmlcorrespondingauthor{Gaojie Jin}{gaojie.jin.kim@gmail.com}

\icmlkeywords{Machine Learning, ICML}

\vskip 0.3in
]



\printAffiliationsAndNotice{} % otherwise use the standard text.

\begin{abstract}
Recent studies have identified a critical challenge in deep neural networks (DNNs) known as ``robust fairness", where models exhibit significant disparities in robust accuracy across different classes. 
While prior work has attempted to address this issue in adversarial robustness, the study of worst-class certified robustness for smoothed classifiers remains unexplored. 
Our work bridges this gap by developing a PAC-Bayesian bound for the worst-class error of smoothed classifiers. 
Through theoretical analysis, we demonstrate that the largest eigenvalue of the smoothed confusion matrix fundamentally influences the worst-class error of smoothed classifiers. 
Based on this insight, we introduce a regularization method that optimizes the largest eigenvalue of smoothed confusion matrix to enhance worst-class accuracy of the smoothed classifier and further improve its worst-class certified robustness. 
We provide extensive experimental validation across multiple datasets and model architectures to demonstrate the effectiveness of our approach.
\end{abstract}

\section{Introduction}

Recent years have witnessed significant advances in improving the adversarial robustness of DNNs \citep{papernot2016distillation,tramer2017ensemble,xu2017feature,madry2017towards,athalye2018obfuscated,wu2020skip}. 
However, models previously considered robust have often been compromised by more sophisticated adversarial attacks \citep{athalye2018obfuscated,uesato2018adversarial,croce2020reliable}. 
This vulnerability has spurred research into certification methods that provide provable guarantees of model robustness within specified perturbation bounds. While notable progress has been made in developing certified robustness techniques for DNNs \citep{katz2017reluplex,wong2018provable,wong2018scaling,huang2019achieving,jia2019certified}, these methods typically require detailed knowledge of model architectures and face scalability challenges across different network designs.

Randomized smoothing \citep{lecuyer2019certified,cohen2019certified,li2019certified} has emerged as a promising solution to these challenges, offering a novel approach for certifying the robustness of smoothed classifiers. 
This method operates by adding smoothing noise to input data and determining predictions based on the most probable label from the smoothed classifier. 
The resulting certification provides guaranteed bounds on the robust radius. 
A key advantage of randomized smoothing lies in its model-agnostic nature, offering efficient certification across diverse architectures without requiring knowledge of their internal structure.

Recent research has highlighted a critical challenge known as ``robust fairness" \citep{xu2021robust,li2023wat,zhang2024towards}, where models exhibit significant disparities in robust accuracy across different classes. 
As demonstrated by \citet{xu2021robust}, the relationship between robustness and fairness is complex, with models often showing substantial variations in their ability to maintain accuracy under adversarial attacks. 
While this issue has received increasing attention in empirical robustness, the fairness aspects of certified robustness remain largely unexplored. Our work addresses this gap by examining fairness specifically in the context of certified robustness, with a focus on enhancing the worst-class performance of smoothed classifiers.

In this work, we develop a PAC-Bayesian bound for the worst-class error of smoothed classifiers. 
Building on the PAC-Bayesian framework introduced by \citet{mcallester1999pac}, which provides probably approximately correct (PAC) guarantees for ``Bayesian-like" learning algorithms, we extend existing theoretical results in two key steps. 
First, we generalize the confusion matrix bounds established by \citet{morvant2012pac} for Gibbs classifiers to the smoothed classifier setting. Following an approach similar to \citet{neyshabur2017pac}, we incorporate the structural information of the base model through a novel chain derivation over confusion matrices. 
Second, we establish a tight relationship between the worst-class error and the largest eigenvalue of confusion matrix, demonstrating that these quantities are connected by a coefficient that approaches one. 
This theoretical development enables us to derive the bound on the worst-class error of smoothed classifiers.

Our theoretical analysis reveals that the worst-class error of smoothed classifiers is bounded by two fundamental components: the principal eigenvalue of the confusion matrix and factors related to model architecture and training data. 
While the latter has been extensively studied and optimized through techniques such as weight decay, batch normalization, and weight spectral regularization\citep{yoshida2017spectral,farnia2018generalizable}, we focus on the former by introducing principal eigenvalue regularization in smooth training. 
Specifically, we propose regularizing the the largest eigenvalue of confusion matrix to improve worst-class accuracy and further enhance worst-class certified robustness of smoothed classifiers.

We validate our approach through comprehensive experiments across multiple datasets. 
Our empirical results demonstrate that the proposed method significantly improves the worst-class certified robustness of smoothed classifiers, leading to more uniform and reliable predictions across all classes.
To summarize, the contributions of this work are as follows:

\begin{itemize}
    \item We establish a theoretical framework for smoothed classifiers by extending PAC-Bayesian generalization analysis to randomized smoothing, enabling us to derive bounds on worst-class error. 
    This represents the first PAC-Bayesian analysis characterizing the worst-class error for smoothed classifiers. 
    
    \item Our theoretical analysis motivates a novel principal eigenvalue regularization technique for smooth training, designed to enhance the worst-class performance of smoothed classifiers.
    We validate our approach through extensive experiments on CIFAR-10, and Tiny-ImageNet datasets, demonstrating significant improvements in worst-class certified robustness.
\end{itemize}

\section{Preliminaries}

Let $\X=\{\x\in\mathbb{R}^d \mid \sum_{i=1}^d x_i^2\le B^2 \}$ represent the input space and $\Y=\{1,...,d_y\}$ denote the label set. 
Consider $\St=\{(\x_i,y_i)_{i=1}^m\}$ a training set of $m$ samples drawn i.i.d. from an unknown but fixed distribution $\mathcal{D}$ over $\X \times \Y$.

We study an $n$-layer neural network with $h$ hidden units per layer with ReLU activation function $\act(\cdot)$. 
Denote the network as $f_{\w}:\X\to \Y$, parameterized by the entire model weights $\w$. 
Let $\W_l$ denote the weight matrix of the $l$-th layer and $\w_l$ be the vectorized form (i.e., $\w_l=\vecc(\W_l)$). 
The network can be expressed as:
$$f_{\w}(\x) = \W_{n}\act(\W_{n-1} ...\act(\W_1 \x)...)$$
where $f_\w^{(1)}(\x) = \W_{1}\x$ and $f_\w^{(i)}(\x) = \W_{i}\act(f_\w^{(i-1)}(\x))$ for $i > 1$. 
For simplicity, biases are incorporated into the weight matrices.
For notation, $\|\W_l\|_2$ denotes the spectral norm (largest singular value) of matrix $\W_l$, $\|\W_l\|_F$ denotes its Frobenius norm, and $\|\w_l\|_p$ denotes the $\ell_p$ vector norm.

We define the empirical and expected confusion matrices for $f_\w$ as $\C_\St^{f_\w}$ and $\C_\D^{f_\w}$ respectively, where $(\C_\St^{f_\w})_{ij}:=0$ and $(\C_\D^{f_\w})_{ij}:=0$, for all $i=j$. 
In the case of $i\ne j$, 
\begin{small}
\begin{equation}\nonumber
\begin{aligned}
&(\C_\St^{f_\w})_{ij}:=\sum_{q=1}^m \frac{1}{m_j}\mathbbm{1}(f_\w(\x_q)[i]\ge \max\limits_{i\ne i'} f_\w(\x_q)[i']) \; \mathbbm{1}(y_q=j),
\end{aligned}
\end{equation}
\begin{equation}\nonumber
\begin{aligned}
&(\C_\D^{f_\w})_{ij}:=\mathbb{P}_{(\mathbf{x}, y) \sim \D}(f_\w(\x)[i]\ge \max\limits_{i\ne i'} f_\w(\x)[i'] \mid y=j),
\end{aligned}
\end{equation}
\end{small}where $m_j$ represents the number of training samples in $\St$ with class label $j$, and $\mathbbm{1}[a\le b]$ is an indicator function that evaluates to 1 when $a\le b$ and 0 otherwise.

\textbf{Randomized smoothing} framework constructs a smoothed classifier $\tilde f_{\w,\vv}$ by enhancing a given base classifier with input noise $\mathbf{v}$, where zero mean Gaussian distribution is a widely employed smoothing distribution in the literature~\citep{lecuyer2019certified,cohen2019certified,li2019certified}.
Typically, in these works, the smoothed classifier predicts the class with the highest confidence on the smoothed data, 
\begin{small}
\begin{equation}
\label{eq:smoothed classifier}
\tilde f_{\w,\vv}(\x):=\underset{c\in\mathcal{Y}}{\arg\max} \; 
\E_{\vv} \mathbbm{1} \Big[f_{\w}(\x+\vv)[c] > \max\limits_{j\ne c} f_{\w}(\x+\vv)[j]\Big].
\end{equation}
\end{small}Let $\vv\sim \mathcal{N}(\mathbf{0},\sigma_v^2 \mathbf{I})$, 
suppose $c_a\in \mathcal{Y}$ and $\underline{p_a}, \overline{p_b}\in [0,1]$ satisfy:
\begin{equation}\nonumber
\begin{aligned}
    &\mathbb{P} (\underset{c}{\argmax} f_{\w}(\x+\vv)[c]=c_a)\ge \underline{p_a} \ge \overline{p_b}\\
    &\quad\quad\quad\quad\quad\quad\quad \ge \max_{j\ne c_a} \mathbb{P} ( \underset{c}{\argmax} f_{\w}(\x+\vv)[c]=j ).
\end{aligned}
\end{equation}

Then,  $\tilde f_{\w,\vv}(\x+\epsilon) = c_a$ holds for all  $\|\epsilon\|_2<R$, where
\begin{equation}
    R=\frac{\sigma_v}{2}(\Psi^{-1}(\underline{p_a})-\Psi^{-1}(\overline{p_b})),
\end{equation}
and $\Psi^{-1}(\cdot)$ is the inverse of the standard Gaussian CDF.

We notate the corresponding empirical confusion matrix and expected confusion matrix generated by smoothed classifier are denoted as $\C^{\tilde f_{\w,\vv}}_{\St}$ and $\C^{\tilde f_{\w,\vv}}_{\D}$, respectively.

%\xiaowei{before mixing randomised smoothing and PAC Bayesian together, I feel it might benefit from having a discussion on them, as they both seem to be working on aggregated classifiers: randomised smoothing aggregates the input noise, and PAC bayesian aggregates the weight noise. }

Building upon margin-based generalization analyses for DNNs in prior works \cite{neyshabur2017pac,farnia2018generalizable}, we introduce a \textbf{margin parameter} $\gamma \ge 0$ and define an empirical margin confusion matrix $\C_{\St,\gamma}^{f_\w}$, where $(\C_{\St,\gamma}^{f_\w})_{ij}=0$ for $i=j$. $\forall i\ne j$, $(\C_{\St,\gamma}^{f_\w})_{ij}$ is defined as
\begin{small}
\begin{equation}\nonumber
\begin{aligned}
\sum_{q=1}^m \frac{1}{m_j}\mathbbm{1}(f_\w(\x_q)[y_q]&\le \gamma+ f_\w(\x_q)[i])  \mathbbm{1}(y_q=j)\\
&\quad\quad\quad\quad\quad\quad\mathbbm{1}(\argmax\limits_{i'\ne y_q}f_\w(\x_q)[i']=i).
\end{aligned}
\end{equation}
\end{small}We define the smoothed classifier under margin $\gamma$ as
\begin{small}
\begin{equation}\nonumber
\label{eq:smoothclassifier}
\begin{aligned}
&\tilde f_{\w,\vv}(\x,\gamma):=\underset{c\in\mathcal{Y}}{\arg\max}\\
&\quad\;\begin{cases}
&\!\E_{\vv} \mathbbm{1} \Big[ f_{\w}(\x+\vv)[c] \!>\! \max\limits_{j\ne c} f_{\w}(\x+\vv)[j] + \gamma \Big], \text{if } c=y \\
&\!\E_{\vv} \mathbbm{1} \Big[ f_{\w}(\x+\vv)[c] + \gamma \!>\! \max\limits_{j\ne c} f_{\w}(\x+\vv)[j] \Big], \text{if } c\ne y 
\end{cases}
\end{aligned}
\end{equation}
\end{small}then the empirical margin confusion matrix of the smoothed classifier is defined as: $\forall i\ne j$, 
\begin{equation}
\label{eq:smoothedconfusedmatrix}
(\C_{\St,\gamma}^{\tilde f_{\w,\vv}})_{ij} := \sum_{q=1}^m \frac{1}{m_j}\mathbbm{1}(\tilde f_{\w,\vv}(\x_q,\gamma)=i) \; \mathbbm{1}(y_q=j).
\end{equation}
\emph{Remark.
Following \citet{neyshabur2017pac}, we introduce an auxiliary variable $\gamma$ to facilitate theoretical development, where $\gamma=0$ recovers the standard empirical confusion matrix. 
That means we set $\gamma=0$ in the bound when the classifier processes unseen data, ensuring that it operates independently of the label.} 
%\xiaowei{how about when $\gamma>0$? }

The \textbf{PAC-Bayesian} framework \citep{mcallester1999pac,mcallester2003simplified} provides tight generalization bounds for the Gibbs classifier \citep{laviolette2005pac}, a stochastic classifier defined over a posterior distribution $Q$ on hypothesis space. 
This framework extends to provide generalization guarantees for the associated $Q$-weighted majority vote classifier, which predicts labels based on the most probable output of the Gibbs classifier \citep{lacasse2006pac,germain2015risk,Gaojie2020}. 
Randomized smoothing and PAC-Bayesian theory share a fundamental connection: both provide probabilistic guarantees through different forms of randomization - randomized smoothing uses input perturbations for certified radius guarantees, while PAC-Bayes employs weight distribution for generalization bounds. 
Building on this connection, we develop the first PAC-Bayesian framework for bounding worst-class error in smoothed classifiers.

The PAC-Bayesian bounds are primarily characterized by the Kullback-Leibler (KL) divergence between the posterior distribution $Q$ and prior distribution $P$ over the hypothesis space. 
Following \citet{neyshabur2017pac}, we define our Gibbs classifiers as $f_{\w+\ul}$, where $\w$ represents the deterministic weights and $\ul$ is a random variable that may depend on the training data. 
Under this formulation, we define the expected and empirical (margin) confusion matrices for the Gibbs classifier $f_{\w+\ul}$ as follows:
\begin{equation}\nonumber
    \C_\D^Q=\E_{\ul}\C_\D^{f_{\w+\ul}}; \quad \C_\St^Q=\E_{\ul}\C_\St^{f_{\w+\ul}}; \quad
    \C_{\St,\gamma}^Q=\E_{\ul}\C_{\St,\gamma}^{f_{\w+\ul}}.
\end{equation}

\citet{morvant2012pac} introduces a PAC-Bayesian bound for the generalization risk of the Gibbs classifier, leveraging the confusion matrix of multi-class labels---a framework that we similarly adopt in this study.
\begin{theorem}[\citet{morvant2012pac}]
\label{thm:2.1}
Consider a training dataset $\St$ with $m$ samples drawn from a distribution $\D$ on $\X \times \Y$ with $\Y = \{1, \ldots, d_y\}$. 
Given a learning algorithm (e.g., a classifier) with prior and posterior distributions $P$ and $Q$ (i.e., $\w+\ul$) on the weights respectively, 
for any $\delta > 0$, with probability $1-\delta$ over the draw of training data, we have that 
\begin{small}
\begin{equation}\nonumber
\|\C^Q_\D \|_2-\| \C^Q_\St\|_2 \leq \sqrt{\frac{8 d_y}{m_{min} - 8d_y} \left[ \KL(Q \| P) + \ln \left( \frac{m_{min}}{4\delta} \right) \right]},
\end{equation}
\end{small}where $m_{min}$ represents the minimal number of examples from $\St$ which belong to the same class.
\end{theorem}

\section{Worst-class error bound for smoothed classifiers}

This section establishes a generalization bound for the worst-class error of smoothed neural network with ReLU activations within the PAC-Bayesian framework. 
Our theoretical analysis extends Thm.~\ref{thm:2.1} through two key developments, which we present in the following.

Our first theoretical development extends the PAC-Bayesian generalization bounds for confusion matrices from Gibbs classifiers to smoothed classifiers. 
Prior research in PAC-Bayesian analysis of neural networks, including \citet{langford2002not}, \citet{neyshabur2017exploring}, and \citet{dziugaite2017computing}, has primarily focused on analyzing KL divergence, perturbation errors, and numerical bound evaluations. 
Notable contributions by \citet{neyshabur2017pac} and \citet{farnia2018generalizable} developed margin-based bounds constructed from weight norms by constraining $\mathbb{P}_{\ul}(f_{\w+\ul}(\x)-f_\w(\x))$. 
%\xiaowei{should be loss expectations over distributions, rather than $f$?}. 
Building upon these approaches, we incorporate the margin operator with smoothed network output restrictions. 
We then establish a bound on $\|\C^{\tilde f_{\w,\vv}}_\D \|_2-\|\C^{\tilde f_{\w,\vv}}_{\St,\gamma}\|_2$ through three key techniques: eigendecomposition of the confusion matrix, margin bounds for smoothed classifier, and application of the Perron–Frobenius theorem.
The main technical challenge lies in establishing bounds on the error of the smoothed classifier's majority vote over noisy inputs --- a previously unaddressed problem in the context of smoothed classifiers.

Our second key advancement leverages the relationship between the largest eigenvalue and maximum column sum to extend our bound from overall classification error to worst-class error on the unseen distribution. 
This theoretical development culminates in Thm.~\ref{thm:main}. 
In this step, the primary challenge is establishing a tight bound between overall error and worst-class error for smoothed classifiers --- a transformation whose tightness we demonstrate empirically in Sec. \ref{sec:proof}. 
We present the complete proof and detailed analysis in Sec.~\ref{sec:proof}.

%\xiaowei{a  discussion on how the presence of input noise in smoothed classifier lead to any technical challenges/differences would help boost the contribution of the paper.  }

\emph{Remark. 
We develop our generalization bound for smoothed classifiers by extending prior PAC-Bayesian analyses through a chain of confusion matrix derivations, building on the foundational work of \citet{morvant2012pac,neyshabur2017pac}. 
This research presents the first PAC-Bayesian generalization bound specifically characterizing the worst-class performance of smoothed classifiers. 
Our extension of the PAC-Bayesian framework to worst-class analysis reveals key factors influencing the worst-class error of smoothed classifiers, providing theoretical foundations for improving certified robustness of the worst-class.}

\begin{theorem}
\label{thm:main}
Consider a training set $\St$ with $m$ samples drawn from a distribution $\D$ over $\X \times\Y$. 
For any $B, n, h > 0$, let the base classifier $f_{\mathbf{w}}: \mathcal{X}_B$ $\rightarrow \mathcal{Y}$ be an $n$-layer feedforward network with $h$ units each layer and ReLU activation function.
Consider $\C_{\St,\gamma}^{\tilde f_{\w,\vv}}$ as the confusion matrix generated by the smoothed classifier $\tilde f_{\w,\vv}$ over $\St$. 
For any $\delta,\gamma>0$, 
with probability at least $1-\delta$, we have 
\begin{small}
\begin{equation}\nonumber
\begin{aligned}
&\max_j \mathop{\mathbb{P}}\limits_{(\mathbf{x}, y) \sim \D}\big(\tilde f_{\w,\vv}(\x)\ne j | y=j\big) - \mu  \lambda_{\max}(\C_{\St,\gamma}^{\tilde f_{\w,\vv}}) \\
&\quad\quad\;\leq\mathcal{O}\left(\sqrt{\frac{\mu^2 d_y}{(m_{min} - 8d_y)\gamma^2} \left[ \Phi(f_\w) + \ln \left( \frac{nm_{min}}{\delta} \right) \right]}\right),
\end{aligned}
\end{equation}
\end{small}where $d_y$ is the number of classes, $m_{min}$ represents the minimal number of examples from $\St$ which belong to the same class, $\Phi(f_\w)=B^2n^2h\ln(nh)\prod_{l=1}^n \|\W_l\|_2^2 \sum_{l=1}^n \frac{\|\W_l\|^2_F}{\|\W_l\|^2_2}$, and $\mu$ is a positive constant which depends on $d_y$.
\end{theorem}

%\xiaowei{Similar as the input noise, it would be helpful to explain how the consideration of worst-class requires any technical challenges. }

\begin{figure}[t!]
\includegraphics[width=0.48
\textwidth]{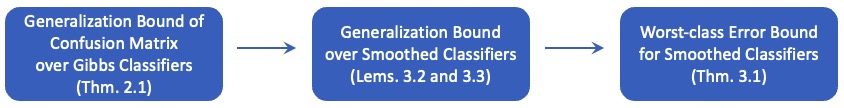}
\centering
\vspace{-5mm}
\caption{
Illustration of the development of Thm.~\ref{thm:main}.
}
\vspace{-3mm}
\label{fig:diagram}
\end{figure}

Thm.~\ref{thm:main} establishes a generalization bound for the worst-class performance of smoothed classifiers through two fundamental components. 
The $\lambda_{\max}(\cdot)$ term is characterized by the largest eigenvalue of the smoothed classifier's empirical confusion matrix, while the final $\mathcal{O}(\cdot)$ term depends on both model architecture and training data characteristics. 
Prior research, including works by \citet{yoshida2017spectral} and \citet{farnia2018generalizable}, has extensively explored methods to optimize the final term through techniques such as weight spectral normalization. 
Our work takes a novel approach by focusing on the $\lambda_{\max}(\cdot)$ term --- the largest eigenvalue of the empirical confusion matrix --- to develop new strategies for enhancing the worst-class certified robustness of smoothed classifiers.

\subsection{Proof for \Cref{thm:main}}
\label{sec:proof}
The first step extends the PAC-Bayesian framework established in Thm.~\ref{thm:2.1}, which provides bounds on the expected confusion matrix of Gibbs classifiers. 
Following the margin-based generalization analysis techniques introduced by \citet{neyshabur2017pac,bartlett2017spectrally}, we develop a novel generalization bound specifically for the smoothed classifier defined in (\ref{eq:smoothed classifier}). 
The details of this bound are presented below.

\begin{lemma}
\label{lem:main1}
Given Thm.~\ref{thm:2.1}, 
let $f_{\mathbf{w}}: \mathcal{X}_B$ $\rightarrow \mathcal{Y}$ denote the base predictor with weights $\mathbf{w}$, and let $P$ be any prior distribution of weights that is independent of the training data, $\w+\ul$ be the posterior of weights over training dataset of size $m$. 
Then, for any $\delta,\gamma>0$, and any random perturbation $\ul,\vv$ $s.t.$ $\pr_{\ul,\vv}(\max_{\x}|f_{\w+\ul}(\x)-f_\w(\x)|_{\infty}<\frac{\gamma}{8} \cap \max_{\x}|f_{\w}(\x+\vv)-f_\w(\x)|_{\infty}<\frac{\gamma}{8})\ge\frac{1}{2}$,
%$\pr_\ul(\max_{\x}|f_{\w+\ul}(\x+\vv)-f_\w(\x)|_{\infty}<\frac{\gamma}{4})\ge\frac{1}{2}$
with probability at least $1-\delta$, we have
\begin{small}
\begin{equation}\nonumber
\begin{aligned}
&\|\C_{\D}^{\tilde f_{\w,\vv}}\|_2 - \|\C_{\St,\gamma}^{\tilde f_{\w,\vv}}\|_2 \leq \\
&\quad\quad\quad 4\sqrt{\frac{8 d_y}{m_{min} - 8d_y} \left[ \KL(\w+\ul \| P) + \ln \left( \frac{3 m_{min}}{4\delta} \right) \right]}.
\end{aligned}
\end{equation}  
\end{small}
\end{lemma}

\textbf{Sketch of Proof.}
\emph{Lem.~\ref{lem:main1} generalizes the bound established in Thm.~\ref{thm:2.1}, extending it from unperturbed input data to the smoothed classifier setting. 
The proof proceeds in two key steps: first, we establish an upper bound on the loss of smoothed classifier through the expected margin loss of the Gibbs classifier; second, we bound this expected loss by the empirical margin loss of the smoothed classifier. 
The complete proof is presented in App.~\ref{app:pf1}.} 
\hfill $\square$

While Lem.~\ref{lem:main1} establishes the relationship between expected and empirical confusion matrices for the smoothed classifier $\tilde f_{\w,\vv}$, its bound involves the KL divergence between posterior and prior distributions. 
Building upon the work of \citet{neyshabur2017pac,farnia2018generalizable}, who developed margin-based bounds dependent on weight norms through sharpness constraints in the PAC-Bayesian framework, we develop a similar approach to replace the KL divergence term. 
However, our analysis requires a dual constraint on both $f_{\w+\ul}(\x)-f_\w(\x)$ and $f_{\w}(\x+\vv)-f_\w(\x)$, as shown in Lem.~\ref{lem:main1}.
This dual constraint arises naturally from our use of smoothed inputs $\x+\vv$.

\begin{lemma}
\label{lem:main2}
Given Lem.~\ref{lem:main1}, for any $B, n, h > 0$, let the base classifier $f_{\mathbf{w}}: \mathcal{X}_B$ $\rightarrow \mathcal{Y}$ be an $n$-layer feedforward network with $h$ units each layer and ReLU activation function.
%Choosing the largest perturbation under the restriction, 
For any $\delta,\gamma>0$, any $\w$ over training dataset of size $m$, with probability at least $1-\delta$, we have the the following bound:
\begin{small}
\begin{equation}\nonumber
\begin{aligned}
&\|\C_{\D}^{\tilde f_{\w,\vv}}\|_2-\|\C_{\St,\gamma}^{\tilde f_{\w,\vv}}\|_2\\
&\quad\;\leq \mathcal{O}\left(\sqrt{\frac{d_y}{(m_{min} - 8d_y)\gamma^2} \left[ \Phi(f_\w) + \ln \left( \frac{nm_{min}}{\delta} \right) \right]}\right),
\end{aligned}
\end{equation}
\end{small}where $\Phi(f_\w)=B^2n^2h\ln(nh)\prod_{l=1}^n ||\W_l||_2^2 \sum_{l=1}^n \frac{||\W_l||^2_F}{||\W_l||^2_2}$.
\end{lemma}

\textbf{Sketch of proof.}
\emph{
The key challenge Lem.~\ref{lem:main2} addressed is computing the KL divergence within the random perturbation bound as established in Lem.~\ref{lem:main1}.
We adopt the pre-determined grid method introduced by \citet{neyshabur2017pac} to construct the prior distribution. 
Through a careful analysis combining both sharpness bounds and Lipschitz continuity properties, we establish an upper bound on the posterior distribution that depends explicitly on the model weight matrices. 
This theoretical framework yields a precise bound on the KL divergence, with the complete derivation provided in App.~\ref{app:pf2}.
}
\hfill $\square$

The above lemma establishes a connection between the spectral norm of the confusion matrix with model weights for smoothed classifiers. 
This connection is bridged through the information encapsulated in the training data and the model weights. 
Our objective, however, extends beyond this result --- we aim to derive an upper bound on the worst-class performance of the smoothed model.

%Since $\|\C_{\D}^{\tilde f_{\w,\vv}}\|_2 - \|\C_{\St,\gamma}^{\tilde f_{\w,\vv}}\|_2 \le \|\C_{\D}^{\tilde f_{\w,\vv}}-\C_{\St,\gamma}^{\tilde f_{\w,\vv}}\|_2$,
%we can easily separate the expected and empirical confusion matrices in the above lemma.
%By leveraging the well-established relationship between the largest column sum and the spectral norm (the largest eigenvalue) of matrices, we can translate the spectral norm bound obtained in the lemma into a bound on the worst-class error of the smoothed model. 
%Specifically, for a general confusion matrix $\C \in \mathbb{R}^{d_y \times d_y}$, we have the following inequality: $\|\C^{\tilde f_{\w,\vv}}_{\D} \|_1 \leq \mu \|\C^{\tilde f_{\w,\vv}}_{\D} \|_2$, where $\mu$ is a constant that depends on the number of classes $d_y$ and is upper bounded by $\sqrt{d_y}$, $\|\C^{\tilde f_{\w,\vv}}_{\D} \|_1$ represents the largest column sum.

%The triangle inequality allows us to bound the difference in spectral norms of the expected and empirical confusion matrices: $\|\C_{\D}^{\tilde f_{\w,\vv}}\|_2 - \|\C_{\St,\gamma}^{\tilde f_{\w,\vv}}\|_2 \le \|\C_{\D}^{\tilde f_{\w,\vv}}-\C_{\St,\gamma}^{\tilde f_{\w,\vv}}\|_2$. 
%This decomposition enables us to analyze the expected and empirical confusion matrices separately in the preceding lemma. 

To bridge the gap between spectral norm bounds and classification performance, we exploit the relationship between a matrix's spectral norm (its largest eigenvalue) and its maximum column sum. 
For a confusion matrix $\C \in \mathbb{R}^{d_y \times d_y}$, this relationship is formalized as $\|\C^{\tilde f_{\w,\vv}}_{\D} \|_1 \leq \mu \lambda_{\max}(\C_{\D}^{\tilde f_{\w,\vv}})$, where $\|\C^{\tilde f{\w,\vv}}_{\D} \|_1$ denotes the maximum column sum, $\lambda_{\max}$ represents the largest eigenvalue, and $\mu$ is a class-dependent constant bounded above by $\sqrt{d_y}$. 
This inequality allows us to translate our spectral norm bound into a constraint on the worst-class error of the smoothed classifier. 
That is, we can get the bound of 
$$\max_j \mathop{\mathbb{P}}\limits_{(\mathbf{x}, y) \sim \D}\big(\tilde f_{\w,\vv}(\x)\ne j | y=j\big) - \mu  \lambda_{\max}(\C_{\St,\gamma}^{\tilde f_{\w,\vv}})$$
in Thm.~\ref{thm:main} through the above inequality and Lem.~\ref{lem:main2}.

\begin{figure}[t!]
\includegraphics[width=0.35
\textwidth]{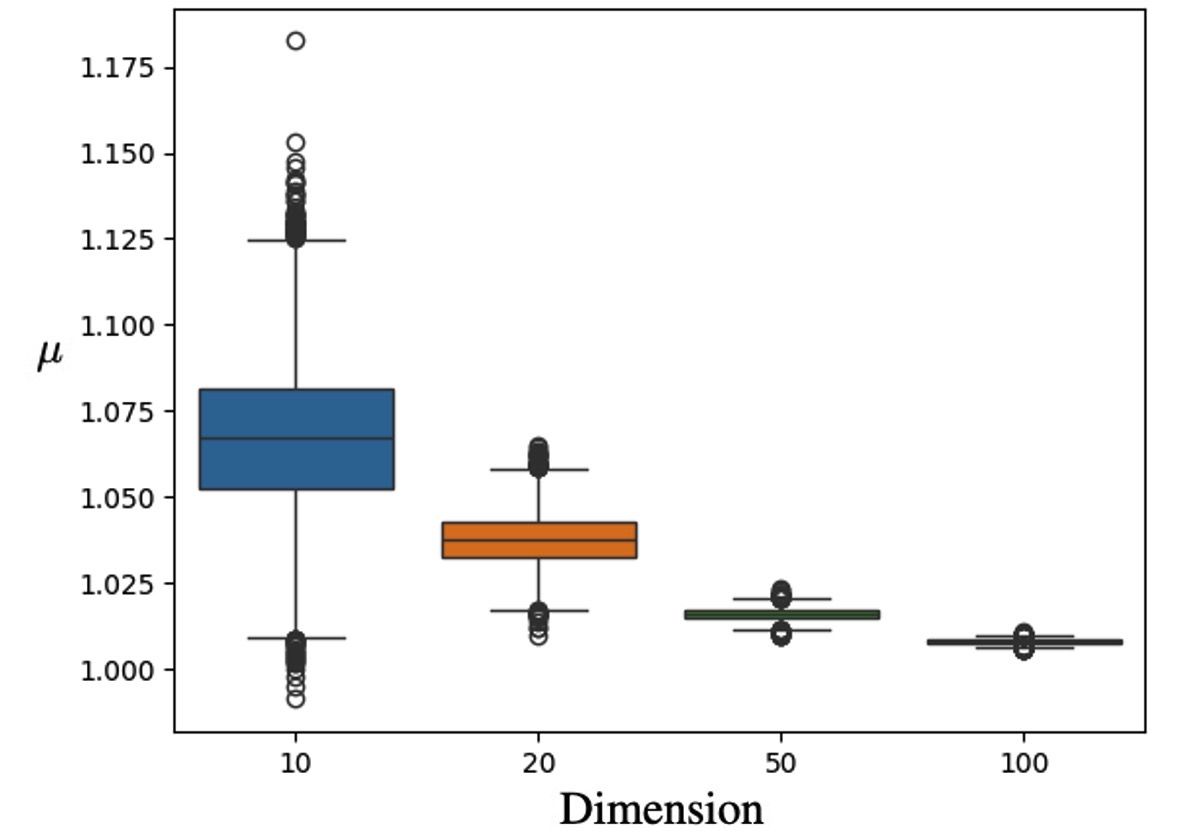}
\centering
\vspace{-3mm}
\caption{
Simulation of $\mu$ under different number of classes. 
Each box (with dimension 10, 20, 50, and 100) is computed by $10000$ randomly generated confusion matrices. 
}
\vspace{-1mm}
\label{fig:box}
\end{figure}

\textbf{Tightness of $\mu$.} We empirically investigate the tightness of our theoretical bound over $\mu$ through extensive numerical experiments, conducting 10,000 simulations with randomly generated confusion matrices across multiple dimensions ($d_y = 10, 20, 50, 100$). 
The results, visualized in Fig.~\ref{fig:box}, demonstrate that the empirical values of $\mu$ consistently approach one across all dimensions. 
Notably, as the dimension increases, we observe a systematic convergence of $\mu$ toward one, exhibiting reduced variance in higher dimensions. 
These numerical findings substantiate the practical significance of our theoretical bounds and confirm their effectiveness in characterizing worst-class error of smoothed classifiers within the PAC-Bayesian framework.

\section{Principal eigenvalue regularization}

Prior works by \citet{cohen2019certified} and \citet{salman2019provably} proposed that incorporating Gaussian noise during training (smooth training) enhances the certified robustness of smoothed classifiers. 
However, the fairness issue of certified robustness, specifically the worst-class performance, remains unconsidered during smooth training. 
Our theoretical analysis in Thm.~\ref{thm:main} demonstrates that the largest eigenvalue of the confusion matrix plays a crucial role in determining the worst-class performance of smoothed classifiers. 
Building upon this insight, we propose a novel enhancement to existing smooth training techniques that explicitly optimizes the confusion matrix's spectral properties, thereby achieving more uniform certified robustness across all classes.

We propose a two-step approach to regularize the largest eigenvalue of the confusion matrix during smooth training. 
The \textbf{first step} focuses on eigenvalue optimization through singular value decomposition (SVD): we calculate ${\partial \lambda_{\max}(\C_{\St,\gamma}^{\tilde f_{\w,\vv}})}/{\partial (\C^{\tilde f_{\w,\vv}}_{\St,\gamma})_{ij}}$, which represents the gradient (sensitivity) of the largest eigenvalue with respect to each element of the confusion matrix.
For time-efficiency, we use 100 samples of $\vv$ to generate $\C_{\St,\gamma}^{\tilde f_{\w,\vv}}$ in our experiments.

This computation relies on a fundamental relationship between a matrix's largest eigenvalue and its SVD.
Specifically, for the confusion matrix $\C^{\tilde f_{\w,\vv}}_{\St,\gamma}$ with SVD:
\begin{equation}
\C^{\tilde f_{\w,\vv}}_{\St,\gamma} = \mathbf{\hat U} \hat \Sigma \mathbf{\hat V}^\top.
\end{equation}
The gradient of the largest eigenvalue is given by:
\begin{equation}
\label{eq:gradient coefficient}
\mathcal{G}_{ij}=\frac{\partial \lambda_{\max}(\C_{\St,\gamma}^{\tilde f_{\w,\vv}})}{\partial (\C^{\tilde f_{\w,\vv}}_{\St,\gamma})_{ij}}=(\mathbf{\hat u}\mathbf{\hat v}^\top)_{ij},
\end{equation}
where $\mathbf{\hat u}$ is the left singular vector corresponding to the largest singular value, $\mathbf{\hat v}$ is the right singular vector corresponding to the largest singular value, and $\mathcal{G}$ represents the matrix of gradients.

\begin{algorithm}[tb]
   \caption{Principal eigenvalue regularization in smooth training}
   \label{alg:example}
\begin{algorithmic}
    \STATE {\bfseries Input:} training set $\St$, network architecture parametrized by $\w$, hyper-parameter $\gamma$, zero mean Gaussian $\mathbf{v}$.
   \STATE {\bfseries Output:} Smoothed classifier $\tilde f_{\w,\vv}$.
   \STATE Randomly initialize network.
   \FOR{each epoch}
   \STATE Generate $\C^{\tilde f_{\w,\vv}}_{\St,\gamma}$ over the training set $\St$ through (\ref{eq:smoothedconfusedmatrix}).
   \STATE Generate the gradient coefficient matrix $\G$ through (\ref{eq:gradient coefficient}).
   \FOR{each minibatch}
   %\STATE Generate adversarial example $\x'$ for $\x$ 
   \STATE Optimize $\mathcal{L}(f_\w(\x+\vv,y))+$
   \STATE $\quad\quad\quad\quad\quad{\color{blue(pigment)}\G_{ij}\cdot \KL(f_\w(\x+\vv)+\gamma(\mathbf{1}-\y)\| \y)}$
   \ENDFOR
   \ENDFOR \\
   Note: $i, j$ depend on the position of $(\x,y)$ in $\C^{\tilde f_{\w,\vv}}_{\St,\gamma}$.
\end{algorithmic}
\end{algorithm}

%In the \textbf{second step}, we estimate the gradient of $\C^{\tilde f_{\w,\vv}}_{\St,\gamma}$ with respect to model weights $\w$.
%The direct computing of the gradient presents significant challenges due to the discrete nature of the elements in $\C^{\tilde f_{\w,\vv}}_{\St,\gamma}$, which are binary $\{0,1\}$ error indicators.
%The non-differentiable nature of these discrete elements precludes the straightforward application of gradient-based optimization techniques.
%Consequently, we are limited to employing computationally expensive methods to estimate discrete gradients, which can significantly hinder the efficiency of the regularizer.

%In the \textbf{second step}, we optimize the empirical confusion matrix $\C^{\tilde f_{\w,\vv}}_{\St,\gamma}$ with respect to the model parameters $\w$. 
%This computation presents a fundamental challenge: the elements of $\C^{\tilde f_{\w,\vv}}_{\St,\gamma}$ are binary indicators in $\{0,1\}$, making them inherently non-differentiable. 
%This discrete nature of the confusion matrix elements prevents the direct application of standard gradient-based optimization methods. 
%As a result, we must resort to discrete gradient estimation techniques, which introduce significant computational overhead and reduce the practical efficiency of the regularization approach.

The \textbf{second step} addresses the optimization problem of model weights $\w$ with respect to the confusion matrix $\C^{\tilde f_{\w,\vv}}_{\St,\gamma}$. 
This optimization faces a fundamental challenge: the confusion matrix consists of binary indicators in $\{0,1\}$ that represent classification errors, making it non-differentiable with respect to model weights. 
This non-differentiability precludes the direct application of conventional gradient-based optimization techniques. 
One can employ discrete gradient estimation methods to deal with this issue, but these introduce substantial computational overhead and impact the efficiency of the regularization approach.

To overcome the non-differentiability challenge, we introduce a surrogate confusion matrix with differentiable elements. 
This matrix preserves the essential structural properties of $\C^{\tilde f_{\w,\vv}}_{\St,\gamma}$, particularly its zero diagonal elements, while enabling gradient-based optimization through using KL divergence rather than 0-1 error.
Specifically, $\forall (\x,y)\in \St_{ij}$, the 0-1 error in $(\C^{\tilde f_{\w,\vv}}_{\St,\gamma})_{ij}$ is replaced by the KL divergence:
\begin{equation}
\label{eq:666}
\KL(f_\w(\x+\vv)+\gamma(\mathbf{1}-\y)\| \y),
\end{equation}
where $\y$ is the one-hot vector of $y$, $\St_{ij}$ represents the sub-set of training samples corresponding to $(\C^{\tilde f_{\w,\vv}}_{\St,\gamma})_{ij}$, $\vv$ is the input smoothing noise. 
Then, combine the first step and second step, the regularization term for all $(\x,y)\in \St_{ij}$ is designed as
\begin{equation}
\label{eq:optimizeweight}
\G_{ij}\cdot\KL(f_\w(\x+\vv)+\gamma(\mathbf{1}-\y)\| \y).
\end{equation}
In (\ref{eq:optimizeweight}), $\G_{ij}$ represents the gradient of the largest eigenvalue with respect to the $(i,j)$-th element of the confusion matrix, while the KL divergence term captures the gradient of the confusion matrix element with respect to model weights.
%and $m_j$ is the number of samples with label $j$ in the training set.

The above optimization strategies draw upon well-established methods. 
First, as shown in (\ref{eq:666}), following smooth training, we optimize the base model using noise-augmented training data rather than directly training the smoothed classifier.
Second, we replace the non-differentiable 0-1 error with differentiable surrogate losses --- specifically KL divergence or cross entropy --- following standard practice in classification optimization. 
Alg.~\ref{alg:example} presents the complete pseudo-code for implementing principal eigenvalue regularization within the smooth training framework, where $\Lc(\cdot)$ represents the standard training loss.

\section{Experiments}

In this section, we present our experimental evaluation, beginning with implementation details and followed by our main results.

\textbf{Datasets.} 
We evaluate our method on two datasets: CIFAR-10 and Tiny-ImageNet. 
CIFAR-10 contains 32×32 pixel images across 10 categories, while Tiny-ImageNet consists of 64×64 pixel images spanning 200 categories. 
Our evaluation protocol uses 500 test images from CIFAR-10, balanced across classes (50 images per class). 
For Tiny-ImageNet, we use 500 test images to evaluate overall certified accuracy. 
To assess worst-class performance, we first identify the worst-class using the original performance of the base model, then evaluate the corresponding certified accuracy using 50 test images from that class.

\textbf{Randomized Smoothing Setting.} 
We configure randomized smoothing with different sampling parameters for each dataset: $N=100,000$ Monte Carlo samples per instance for CIFAR-10 and $N=10,000$ for Tiny-ImageNet. 
Certified robustness is evaluated across multiple noise levels: $\sigma_v\in\{0.12, 0.25, 0.5\}$ for CIFAR-10 and $\sigma_v\in\{0.05, 0.15, 0.25\}$ for Tiny-ImageNet. 
For each test example, we calculate the certified radius at these noise levels and compute the proportion of examples whose radius exceeds specified thresholds. 
For CIFAR-10, we report certified accuracy at each radius for individual noise levels, while for Tiny-ImageNet, we report the best certified accuracy across all noise levels at each radius.

\begin{table*}[t!]
\centering
%\captionsetup{format=myformat}
\caption{Certified accuracy of CIFAR-10 smoothed classifiers at various $\ell_2$ radii.}
\label{tab:1}
\vspace{1mm}
\renewcommand\arraystretch{1.35}
\scalebox{0.8}{
\begin{tabular}{cclcccccccccccc}
\specialrule{.1em}{.075em}{.075em} 
Smoothing && \multicolumn{1}{c}{$\ell_2$ radius} && \multicolumn{2}{c}{0.12} && \multicolumn{2}{c}{0.25} && \multicolumn{2}{c}{0.5} && \multicolumn{2}{c}{1.0} \\ 
Noise && \multicolumn{1}{c}{Accuracy (\%)} && Overall & Worst && Overall & Worst && Overall & Worst && Overall & Worst   \\ 
\cline{1-1} \cline{3-3} \cline{5-6} \cline{8-9} \cline{11-12} \cline{14-15}
\multirow{9}{*}{$\sigma=0.12$} && Baseline~\citep{cohen2019certified}  && 71.4 & 38 && 58.6 & 26 && 0. & 0. && 0. & 0.  \\
&& \;\; + \citet{xu2021robust} && 70.6 & 40 && 57.0 & 26 && 0. & 0. && 0. & 0.   \\
&& \;\; + \citet{wei2023cfa} && 70.2 & 42 && 56.6 & 28 && 0. & 0. && 0. & 0.  \\
&& \;\; + \citet{li2023wat} && 71.0 & 40 && 57.4 & 28 && 0. & 0. && 0. & 0. \\
&& \;\; + \citet{zhang2024towards} && 71.4 & 44 && 57.8 & \bf 30 && 0. & 0. && 0. & 0.  \\
&& \grr \;\; + Ours &\grr & \grr \bf 71.8 & \grr \bf 46 &\grr & \grr \bf 59.4 & \grr \bf 30 &\grr & \grr 0. & \grr 0. &\grr & \grr 0. & \grr 0.  \\
\cline{3-3} \cline{5-6} \cline{8-9} \cline{11-12} \cline{14-15}
&& Baseline~\citep{salman2019provably}  && \bf 75.4 & 46 && 69.0 & 26 && 0. & 0. && 0. & 0.   \\
&& \;\; + \citet{zhang2024towards} && 74.2 & 48 && 67.6 & 30 && 0. & 0. && 0. & 0.  \\
&& \grr \;\; + Ours &\grr & \grr \bf 75.4 & \grr \textbf{50} &\grr & \grr \bf 69.2 & \grr \bf 32 &\grr & \grr 0. & \grr 0. &\grr & \grr 0. & \grr 0.   \\
\hline
\multirow{9}{*}{$\sigma=0.25$} && Baseline~\citep{cohen2019certified}  && 68.0 & 42 && 60.0 & 32 && 42.8 & 20 && 0. & 0.  \\
&& \;\; + \citet{xu2021robust} && 67.6 & 42 && 59.4 & 34 && 42.4 & 20 && 0. & 0.  \\
&& \;\; + \citet{wei2023cfa} && 68.2 & 44 && \bf 60.6 & 32 && 43.2 & 22 && 0. & 0.   \\
&& \;\; + \citet{li2023wat} && 67.0 & 46 && 59.2 & 34 && 42.0 & \bf 24 && 0. & 0.  \\
&& \;\; + \citet{zhang2024towards} && 68.4 & 44 && 60.2 & 36 && 43.0 & 22 && 0. & 0.  \\
&& \grr \;\; + Ours &\grr & \grr \bf 69.2 & \grr \bf 48 &\grr & \grr 60.0 & \grr \bf 42 &\grr & \grr \bf 43.8 & \grr \bf 24 &\grr & \grr 0. & \grr 0.  \\
\cline{3-3} \cline{5-6} \cline{8-9} \cline{11-12} \cline{14-15}
&& Baseline~\citep{salman2019provably}  && \bf 69.4 & 42 && \bf 63.4 & 36 && \bf50.0 & 22 && 0. & 0.   \\
&& \;\; + \citet{zhang2024towards} && 69.0 & 46 && 62.8 & 40 && 49.2 & 24 && 0. & 0.  \\
&& \grr \;\; + Ours &\grr & \grr \bf 69.4 & \grr \bf 50 &\grr & \grr 63.2 & \grr \bf 42 &\grr & \grr \bf 50.0 & \grr \bf 28 &\grr & \grr 0. & \grr 0.  \\
\hline
\multirow{9}{*}{$\sigma=0.5$} && Baseline~\citep{cohen2019certified}  && 59.4 & 36 && \bf 54.6 & 26 && \bf 41.4 & 14 && 23.4 & 2  \\
&& \;\; + \citet{xu2021robust} && 58.8 & 38 && 54.2 & 30 && 41.4 & 16 && 22.6 & 2  \\
&& \;\; + \citet{wei2023cfa} && 58.0 & 40 && 53.6 & 30 && 40.8 & 18 && 22.0 & 2   \\
&& \;\; + \citet{li2023wat} && 59.2 & 40 && 54.4 & 32 && 41.0 & 18 && 23.0 & \bf 4  \\
&& \;\; + \citet{zhang2024towards} && 58.6 & 38 && 53.2 & 32 && 40.2 & 16 && 22.2 &  2  \\
&& \grr \;\; + Ours &\grr & \grr \bf 59.8 & \grr \bf 42 &\grr & \grr \bf 54.6 & \grr \bf 34 &\grr & \grr 41.2 & \grr \bf 20 &\grr & \grr \bf 23.6 & \grr \bf 4  \\
\cline{3-3} \cline{5-6} \cline{8-9} \cline{11-12} \cline{14-15}
&& Baseline~\citep{salman2019provably}  && 59.6 & 26 && 54.0 & 24 && \bf 45.4 & 12 && \bf 31.0 & 2   \\
&& \;\; + \citet{zhang2024towards} && 58.0 & 30 && 52.8 & 28 && 43.6 & 14 && 30.2 & 2  \\
&& \grr \;\; + Ours &\grr & \grr \bf 59.8 & \grr \bf 38 &\grr & \grr \bf 54.4 & \grr \bf 32 &\grr & \grr 45.0 & \grr \bf 18 &\grr & \grr \bf 31.0 & \grr \bf 4   \\
\specialrule{.1em}{.075em}{.075em} 
\end{tabular}
}
\vspace{0mm}
\end{table*}

\textbf{Baselines for CIFAR-10.} 
We evaluate our method on ResNet-110 \citep{he2016deep} against current state-of-the-art approaches for robust fairness: FRL \citep{xu2021robust}, FAAL \citep{zhang2024towards}, CFA \citep{wei2023cfa}, and WAT \citep{li2023wat}. 
FRL builds upon TRADES \citep{zhang2019theoretically}, implementing two key strategies: reweight and remargin. Following \citet{zhang2024towards}, we use their most effective variant, FRL-RWRM, with fairness constraint parameters $\tau_1 = \tau_2 = 0.07$ for both reweight and remargin components. 
For FAAL, we implement its strongest version, which incorporates AWP \citep{DBLP:conf/nips/WuX020}. 
All methods are evaluated by fine-tuning baseline models from \citet{cohen2019certified} and \citet{salman2019provably} for 10 epochs, with a learning rate of 0.001, momentum of 0.9, batch size of 256, and weight decay of $5\times 10^{-4}$.
We set $\gamma=0.1$ for our fine-tuning method.

\textbf{Baselines for Tiny-ImageNet.} 
The Diffusion Probabilistic Model (DDPM) \citep{ho2020denoising,nichol2021improved,carlini2022certified} is a sophisticated generative framework that produces high-quality samples by learning to reverse a gradual noise addition process. 
In our experiments for Tiny-ImageNet, we use a baseline model with WideResNet-28-10 architecture \citep{DBLP:conf/bmvc/ZagoruykoK16} trained on 50M DDPM-generated samples following \citet{wang2023better}.
Our method is used to fine-tune the pre-trained model within 5 epochs on the standard dataset.
Our fine-tuning setting is the same as in the previous CIFAR-10 experiment.

\begin{table}[t!]
\centering
%\captionsetup{format=myformat}
\caption{Certified accuracy of Tiny-ImageNet smoothed classifiers at various $\ell_2$ radii.}
\label{tab:2}
\vspace{1mm}
\renewcommand\arraystretch{1.35}
\scalebox{0.7}{
\begin{tabular}{lccccccccc}
\specialrule{.1em}{.075em}{.075em} 
\multicolumn{1}{c}{$\ell_2$ radius} && \multicolumn{2}{c}{0.} && \multicolumn{2}{c}{0.05} && \multicolumn{2}{c}{0.15} \\ 
\multicolumn{1}{c}{Accuracy (\%)} && Overall & Worst && Overall & Worst && Overall & Worst   \\ 
\cline{1-1} \cline{3-4} \cline{6-7} \cline{9-10} 
Baseline  && 64.8 & 26 && 48.6 & 12 && \bf 37.2 & 2   \\
\;\; + Ours && \bf 65.0 & \bf 32 && \bf 49.2 & \bf 18 && \bf 37.2 & \bf 4  \\
\specialrule{.1em}{.075em}{.075em} 
\end{tabular}
}
\vspace{-2mm}
\end{table}

\subsection{Main empirical results}

Tab.~\ref{tab:1} presents our certified accuracy results on CIFAR-10 across smooth noise levels $\sigma \in {0.12, 0.25, 0.5}$. 
Our method consistently outperforms existing approaches (FRL, FAAL, CFA, and WAT), demonstrating higher overall certified accuracy while achieving superior worst-class performance. 
This improvement is particularly evident at $\sigma=0.12$, where our method achieves the best performance across all metrics using both \citet{cohen2019certified} and \citet{salman2019provably} baseline models.

We extend our evaluation to Tiny-ImageNet using the baseline models from \citet{wang2023better}, maintaining the same fine-tuning protocol as our CIFAR-10 experiments. 
For each certification radius, we report the highest certified accuracy achieved across noise levels $\sigma_v\in\{0.05, 0.15, 0.25\}$. 
Tab. \ref{tab:2} shows that after 5 epochs of fine-tuning, our method substantially improves worst-class certified accuracy while preserving overall certified performance.

\begin{figure}[t!]
\includegraphics[width=0.48
\textwidth]{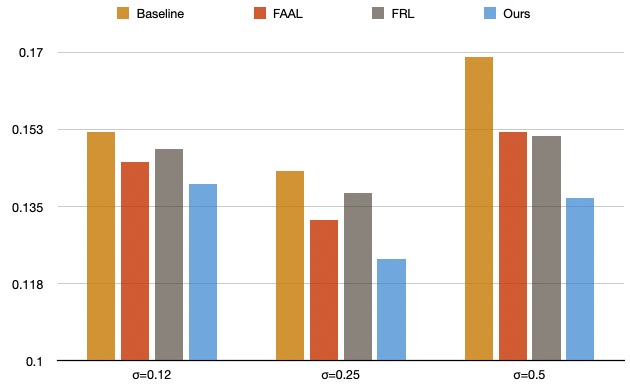}
\centering
\vspace{-8mm}
\caption{
The class-wise variation in certified accuracy at radius 0.12 for CIFAR-10, measured as \textbf{standard deviation} for smooth noise levels $\sigma\in\{0.12,0.25,0.50\}$ respectively.
}
\vspace{-2mm}
\label{fig:std}
\end{figure}

Furthermore, Fig.~\ref{fig:std} compares the standard deviation of class-wise certified accuracy at radius 0.12 across noise levels $\sigma\in\{0.12,0.25,0.5\}$ for different fine-tuning methods applied to the smoothed ResNet-110 model. 
Our method achieves more uniform (lower standard deviation) certified accuracy across classes, demonstrating improved fairness in robustness certification.

\section{Related work}
The development of randomized smoothing \citep{lecuyer2019certified,cohen2019certified,li2019certified} has spawned numerous theoretical and practical advances. Early challenges were identified by \citet{kumar2020curse}, who demonstrated limitations in extending smoothing to various attack models.
Several researchers expanded the scope of randomized smoothing to new domains: \citet{lee2019tight,schuchardt2020collective,wang2021certified} addressed discrete perturbations including $\ell_0$-attacks, while \citet{bojchevski2020efficient,gao2020certified,levine2020randomized,liu2021pointguard} explored applications to graphs, patches, and point clouds. Certification methods for multiple norms ($\ell_1$, $\ell_2$, and $\ell_\infty$) were developed by \citet{yang2020randomized,dinghuaizhang}.
The framework has been adapted to various applications: certified federated learning against backdoor attacks \citep{xie2021crfl}, multi-agent reinforcement learning with certified bounds \citep{mu2023certified}, and defense against data poisoning \citep{rosenfeld2020certified,levine2020deep,jia2021intrinsic,weber2023rab}. Theoretical advancements include \citet{dvijotham2020framework}'s unified treatment of discrete and continuous smoothing, \citet{mohapatra2020higher}'s gradient-enhanced certificates, and \citet{horvath2021boosting}'s ensemble approaches.
Beyond traditional norm-based certificates, the field has expanded to certify geometric transformations \citep{fischer2020certified,li2021tss}, regression and segmentation tasks \citep{chiang2020detection,fischer2021scalable}, and top-k predictions \citep{jia2019certified} and classifier confidence \citep{kumar2020certifying}.
Recent innovations include universal certification frameworks \citep{hong2022unicr}, generalized smoothing for semantic transformations \citep{hao2022gsmooth}, and gray-box approaches incorporating invariance knowledge \citep{schuchardt2022invariance}. Notable challenges and improvements have been identified, including poisoning attacks \citep{mehra2021robust}, input-dependent optimization of Gaussian variance \citep{alfarra2022data}, decision boundary effects \citep{mohapatra2021hidden}, scalability solutions \citep{alfarra2022deformrs}, and efficient certification methods \citep{chen2022input}.

Recent research has focused on addressing fairness in adversarial robustness, specifically the disparity in robust accuracy across different classes \citep{li2023wat,ma2022tradeoff,sun2023improving,wei2023cfa,zhang2024towards}. 
\citet{xu2021robust} first identified that conventional adversarial training can lead to significant disparities in accuracy and robustness across different data groups while improving average robustness. They introduced the Fair-Robust-Learning (FRL) framework, which uses reweight and remargin strategies to fine-tune pre-trained models and reduce boundary errors.
\citet{ma2022tradeoff} identified a trade-off between robustness and robustness fairness, showing that larger perturbation radii in adversarial training lead to increased variance.  
\citet{sun2023improving} developed Balance Adversarial Training (BAT), which dynamically adjusts per-class attack strengths to generate boundary samples for more balanced learning.
Recent advances include more sophisticated approaches: \citet{wei2023cfa} proposed Class-wise Fair Adversarial training (CFA), which automatically tailors training configurations for each class to enhance worst-class robustness while maintaining overall performance. 
\citet{jin2025enhancing} studied the worst-class adversarial robustness performance from the PAC-Bayesian perspective and proposed a spectral regularization method. 
\citet{li2023wat} introduced Worst-class Adversarial Training (WAT), employing no-regret dynamics to optimize worst-class robust risk.

In contrast to previous research, we establish a theoretical bound on worst-class error for smoothed classifiers and introduce a novel approach to enhance worst-class certified accuracy through largest eigenvalue regularization of the smoothed confusion matrix.
To the best of our knowledge, this work presents the first worst-class error bound and corresponding regularization technique for smoothed classifiers within the randomized smoothing framework.

\section{Conclusion}

Our work identifies a critical gap in certified robustness: the need to systematically consider the principal eigenvalue of smoothed classifiers' confusion matrices. 
Through a novel PAC-Bayesian theoretical framework, efficient principal eigenvalue regularization techniques, and comprehensive empirical validation, we demonstrate that optimizing the confusion matrix's principal eigenvalue significantly enhances the worst-class certified robustness of smoothed classifiers and improves the uniformness of certified robust performance across classes.

\clearpage


\section*{Impact Statement}

This paper presents work whose goal is to advance the field of 
Machine Learning. There are many potential societal consequences 
of our work, none which we feel must be specifically highlighted here.

\nocite{langley00}

\bibliography{example_paper}

\begin{thebibliography}{80}
\providecommand{\natexlab}[1]{#1}
\providecommand{\url}[1]{\texttt{#1}}
\expandafter\ifx\csname urlstyle\endcsname\relax
  \providecommand{\doi}[1]{doi: #1}\else
  \providecommand{\doi}{doi: \begingroup \urlstyle{rm}\Url}\fi

\bibitem[Alfarra et~al.(2022{\natexlab{a}})Alfarra, Bibi, Khan, Torr, and Ghanem]{alfarra2022deformrs}
Alfarra, M., Bibi, A., Khan, N., Torr, P.~H., and Ghanem, B.
\newblock Deformrs: Certifying input deformations with randomized smoothing.
\newblock In \emph{AAAI}, 2022{\natexlab{a}}.

\bibitem[Alfarra et~al.(2022{\natexlab{b}})Alfarra, Bibi, Torr, and Ghanem]{alfarra2022data}
Alfarra, M., Bibi, A., Torr, P.~H., and Ghanem, B.
\newblock Data dependent randomized smoothing.
\newblock In \emph{UAI}, 2022{\natexlab{b}}.

\bibitem[Athalye et~al.(2018)Athalye, Carlini, and Wagner]{athalye2018obfuscated}
Athalye, A., Carlini, N., and Wagner, D.
\newblock Obfuscated gradients give a false sense of security: Circumventing defenses to adversarial examples.
\newblock In \emph{ICML}, 2018.

\bibitem[Bandeira \& Boedihardjo(2021)Bandeira and Boedihardjo]{bandeira2021spectral}
Bandeira, A.~S. and Boedihardjo, M.~T.
\newblock The spectral norm of gaussian matrices with correlated entries.
\newblock \emph{arXiv preprint arXiv:2104.02662}, 2021.

\bibitem[Bartlett et~al.(2017)Bartlett, Foster, and Telgarsky]{bartlett2017spectrally}
Bartlett, P.~L., Foster, D.~J., and Telgarsky, M.~J.
\newblock Spectrally-normalized margin bounds for neural networks.
\newblock \emph{Advances in neural information processing systems}, 30, 2017.

\bibitem[Bojchevski et~al.(2020)Bojchevski, Gasteiger, and G{\"u}nnemann]{bojchevski2020efficient}
Bojchevski, A., Gasteiger, J., and G{\"u}nnemann, S.
\newblock Efficient robustness certificates for discrete data: Sparsity-aware randomized smoothing for graphs, images and more.
\newblock In \emph{ICML}, 2020.

\bibitem[Carlini et~al.(2022)Carlini, Tramer, Dvijotham, Rice, Sun, and Kolter]{carlini2022certified}
Carlini, N., Tramer, F., Dvijotham, K.~D., Rice, L., Sun, M., and Kolter, J.~Z.
\newblock (certified!!) adversarial robustness for free!
\newblock \emph{arXiv preprint arXiv:2206.10550}, 2022.

\bibitem[Chen et~al.(2022)Chen, Li, Yan, Li, and Sheng]{chen2022input}
Chen, R., Li, J., Yan, J., Li, P., and Sheng, B.
\newblock Input-specific robustness certification for randomized smoothing.
\newblock In \emph{AAAI}, 2022.

\bibitem[Chiang et~al.(2020)Chiang, Curry, Abdelkader, Kumar, Dickerson, and Goldstein]{chiang2020detection}
Chiang, P.-y., Curry, M., Abdelkader, A., Kumar, A., Dickerson, J., and Goldstein, T.
\newblock Detection as regression: Certified object detection with median smoothing.
\newblock \emph{NeurIPS}, 2020.

\bibitem[Cohen et~al.(2019)Cohen, Rosenfeld, and Kolter]{cohen2019certified}
Cohen, J., Rosenfeld, E., and Kolter, Z.
\newblock Certified adversarial robustness via randomized smoothing.
\newblock In \emph{International conference on machine learning}, pp.\  1310--1320. PMLR, 2019.

\bibitem[Croce \& Hein(2020)Croce and Hein]{croce2020reliable}
Croce, F. and Hein, M.
\newblock Reliable evaluation of adversarial robustness with an ensemble of diverse parameter-free attacks.
\newblock In \emph{International Conference on Machine Learning}, pp.\  2206--2216. PMLR, 2020.

\bibitem[Dvijotham et~al.(2020)Dvijotham, Hayes, Balle, Kolter, Qin, Gyorgy, Xiao, Gowal, and Kohli]{dvijotham2020framework}
Dvijotham, K.~D., Hayes, J., Balle, B., Kolter, Z., Qin, C., Gyorgy, A., Xiao, K., Gowal, S., and Kohli, P.
\newblock A framework for robustness certification of smoothed classifiers using f-divergences.
\newblock \emph{ICLR}, 2020.

\bibitem[Dziugaite \& Roy(2017)Dziugaite and Roy]{dziugaite2017computing}
Dziugaite, G.~K. and Roy, D.~M.
\newblock Computing nonvacuous generalization bounds for deep (stochastic) neural networks with many more parameters than training data.
\newblock \emph{arXiv preprint arXiv:1703.11008}, 2017.

\bibitem[Farnia et~al.(2019)Farnia, Zhang, and Tse]{farnia2018generalizable}
Farnia, F., Zhang, J.~M., and Tse, D.
\newblock Generalizable adversarial training via spectral normalization.
\newblock In \emph{ICLR}, 2019.

\bibitem[Fischer et~al.(2020)Fischer, Baader, and Vechev]{fischer2020certified}
Fischer, M., Baader, M., and Vechev, M.
\newblock Certified defense to image transformations via randomized smoothing.
\newblock \emph{NeurIPS}, 2020.

\bibitem[Fischer et~al.(2021)Fischer, Baader, and Vechev]{fischer2021scalable}
Fischer, M., Baader, M., and Vechev, M.
\newblock Scalable certified segmentation via randomized smoothing.
\newblock In \emph{ICML}, 2021.

\bibitem[Frobenius et~al.(1912)Frobenius, Frobenius, Frobenius, Frobenius, and Mathematician]{frobenius1912matrizen}
Frobenius, G., Frobenius, F.~G., Frobenius, F.~G., Frobenius, F.~G., and Mathematician, G.
\newblock {\"U}ber matrizen aus nicht negativen elementen.
\newblock 1912.

\bibitem[Gao et~al.(2020)Gao, Hu, and Gong]{gao2020certified}
Gao, Z., Hu, R., and Gong, Y.
\newblock Certified robustness of graph classification against topology attack with randomized smoothing.
\newblock In \emph{GLOBECOM 2020-2020 IEEE Global Communications Conference}, 2020.

\bibitem[Germain et~al.(2015)Germain, Lacasse, Laviolette, Marchand, and Roy]{germain2015risk}
Germain, P., Lacasse, A., Laviolette, F., Marchand, M., and Roy, J.-F.
\newblock Risk bounds for the majority vote: From a pac-bayesian analysis to a learning algorithm.
\newblock \emph{arXiv preprint arXiv:1503.08329}, 2015.

\bibitem[Hao et~al.(2022)Hao, Ying, Dong, Su, Song, and Zhu]{hao2022gsmooth}
Hao, Z., Ying, C., Dong, Y., Su, H., Song, J., and Zhu, J.
\newblock Gsmooth: Certified robustness against semantic transformations via generalized randomized smoothing.
\newblock In \emph{ICML}, 2022.

\bibitem[He et~al.(2016)He, Zhang, Ren, and Sun]{he2016deep}
He, K., Zhang, X., Ren, S., and Sun, J.
\newblock Deep residual learning for image recognition.
\newblock In \emph{CVPR}, 2016.

\bibitem[Ho et~al.(2020)Ho, Jain, and Abbeel]{ho2020denoising}
Ho, J., Jain, A., and Abbeel, P.
\newblock Denoising diffusion probabilistic models.
\newblock \emph{Advances in neural information processing systems}, 33:\penalty0 6840--6851, 2020.

\bibitem[Hong et~al.(2022)Hong, Wang, and Hong]{hong2022unicr}
Hong, H., Wang, B., and Hong, Y.
\newblock Unicr: Universally approximated certified robustness via randomized smoothing.
\newblock In \emph{ECCV}. Springer, 2022.

\bibitem[Horv{\'a}th et~al.(2021)Horv{\'a}th, Mueller, Fischer, and Vechev]{horvath2021boosting}
Horv{\'a}th, M.~Z., Mueller, M.~N., Fischer, M., and Vechev, M.
\newblock Boosting randomized smoothing with variance reduced classifiers.
\newblock In \emph{ICLR}, 2021.

\bibitem[Huang et~al.(2019)Huang, Stanforth, Welbl, Dyer, Yogatama, Gowal, Dvijotham, and Kohli]{huang2019achieving}
Huang, P.-S., Stanforth, R., Welbl, J., Dyer, C., Yogatama, D., Gowal, S., Dvijotham, K., and Kohli, P.
\newblock Achieving verified robustness to symbol substitutions via interval bound propagation.
\newblock \emph{arXiv preprint arXiv:1909.01492}, 2019.

\bibitem[Jia et~al.(2020)Jia, Cao, Wang, and Gong]{jia2019certified}
Jia, J., Cao, X., Wang, B., and Gong, N.~Z.
\newblock Certified robustness for top-k predictions against adversarial perturbations via randomized smoothing.
\newblock In \emph{ICLR}, 2020.

\bibitem[Jia et~al.(2021)Jia, Cao, and Gong]{jia2021intrinsic}
Jia, J., Cao, X., and Gong, N.~Z.
\newblock Intrinsic certified robustness of bagging against data poisoning attacks.
\newblock In \emph{AAAI}, 2021.

\bibitem[Jin et~al.(2020)Jin, Yi, Zhang, Zhang, Schewe, and Huang]{Gaojie2020}
Jin, G., Yi, X., Zhang, L., Zhang, L., Schewe, S., and Huang, X.
\newblock How does weight correlation affect the generalisation ability of deep neural networks.
\newblock \emph{NeurIPS}, 2020.

\bibitem[Jin et~al.(2025)Jin, Wu, Liu, Huang, and Mu]{jin2025enhancing}
Jin, G., Wu, S., Liu, J., Huang, T., and Mu, R.
\newblock Enhancing robust fairness via confusional spectral regularization.
\newblock \emph{arXiv preprint arXiv:2501.13273}, 2025.

\bibitem[Katz et~al.(2017)Katz, Barrett, Dill, Julian, and Kochenderfer]{katz2017reluplex}
Katz, G., Barrett, C., Dill, D.~L., Julian, K., and Kochenderfer, M.~J.
\newblock Reluplex: An efficient smt solver for verifying deep neural networks.
\newblock In \emph{International Conference on Computer Aided Verification}, pp.\  97--117. Springer, 2017.

\bibitem[Kumar et~al.(2020{\natexlab{a}})Kumar, Levine, Feizi, and Goldstein]{kumar2020certifying}
Kumar, A., Levine, A., Feizi, S., and Goldstein, T.
\newblock Certifying confidence via randomized smoothing.
\newblock \emph{NeurIPS}, 2020{\natexlab{a}}.

\bibitem[Kumar et~al.(2020{\natexlab{b}})Kumar, Levine, Goldstein, and Feizi]{kumar2020curse}
Kumar, A., Levine, A., Goldstein, T., and Feizi, S.
\newblock Curse of dimensionality on randomized smoothing for certifiable robustness.
\newblock In \emph{ICML}, 2020{\natexlab{b}}.

\bibitem[Lacasse et~al.(2006)Lacasse, Laviolette, Marchand, Germain, and Usunier]{lacasse2006pac}
Lacasse, A., Laviolette, F., Marchand, M., Germain, P., and Usunier, N.
\newblock Pac-bayes bounds for the risk of the majority vote and the variance of the gibbs classifier.
\newblock \emph{Advances in Neural information processing systems}, 19, 2006.

\bibitem[Langford \& Caruana(2002)Langford and Caruana]{langford2002not}
Langford, J. and Caruana, R.
\newblock (not) bounding the true error.
\newblock \emph{Advances in Neural Information Processing Systems}, 2:\penalty0 809--816, 2002.

\bibitem[Laviolette \& Marchand(2005)Laviolette and Marchand]{laviolette2005pac}
Laviolette, F. and Marchand, M.
\newblock Pac-bayes risk bounds for sample-compressed gibbs classifiers.
\newblock In \emph{Proceedings of the 22nd international conference on Machine learning}, 2005.

\bibitem[Lecuyer et~al.(2019)Lecuyer, Atlidakis, Geambasu, Hsu, and Jana]{lecuyer2019certified}
Lecuyer, M., Atlidakis, V., Geambasu, R., Hsu, D., and Jana, S.
\newblock Certified robustness to adversarial examples with differential privacy.
\newblock In \emph{2019 IEEE Symposium on Security and Privacy (SP)}, pp.\  656--672. IEEE, 2019.

\bibitem[Lee et~al.(2019)Lee, Yuan, Chang, and Jaakkola]{lee2019tight}
Lee, G.-H., Yuan, Y., Chang, S., and Jaakkola, T.
\newblock Tight certificates of adversarial robustness for randomly smoothed classifiers.
\newblock \emph{NeurIPS}, 2019.

\bibitem[Levine \& Feizi(2020{\natexlab{a}})Levine and Feizi]{levine2020deep}
Levine, A. and Feizi, S.
\newblock Deep partition aggregation: Provable defenses against general poisoning attacks.
\newblock In \emph{ICLR}, 2020{\natexlab{a}}.

\bibitem[Levine \& Feizi(2020{\natexlab{b}})Levine and Feizi]{levine2020randomized}
Levine, A. and Feizi, S.
\newblock (de) randomized smoothing for certifiable defense against patch attacks.
\newblock \emph{NeurIPS}, 2020{\natexlab{b}}.

\bibitem[Li \& Liu(2023)Li and Liu]{li2023wat}
Li, B. and Liu, W.
\newblock Wat: improve the worst-class robustness in adversarial training.
\newblock In \emph{Proceedings of the AAAI Conference on Artificial Intelligence}, 2023.

\bibitem[Li et~al.(2019)Li, Chen, Wang, and Carin]{li2019certified}
Li, B., Chen, C., Wang, W., and Carin, L.
\newblock Certified adversarial robustness with additive noise.
\newblock \emph{Advances in neural information processing systems}, 32, 2019.

\bibitem[Li et~al.(2021)Li, Weber, Xu, Rimanic, Kailkhura, Xie, Zhang, and Li]{li2021tss}
Li, L., Weber, M., Xu, X., Rimanic, L., Kailkhura, B., Xie, T., Zhang, C., and Li, B.
\newblock Tss: Transformation-specific smoothing for robustness certification.
\newblock In \emph{Proceedings of the 2021 ACM SIGSAC Conference on Computer and Communications Security}, 2021.

\bibitem[Liu et~al.(2021)Liu, Jia, and Gong]{liu2021pointguard}
Liu, H., Jia, J., and Gong, N.~Z.
\newblock Pointguard: Provably robust 3d point cloud classification.
\newblock In \emph{CVPR}, 2021.

\bibitem[Ma et~al.(2022)Ma, Wang, and Liu]{ma2022tradeoff}
Ma, X., Wang, Z., and Liu, W.
\newblock On the tradeoff between robustness and fairness.
\newblock \emph{Advances in Neural Information Processing Systems}, 35:\penalty0 26230--26241, 2022.

\bibitem[Madry et~al.(2017)Madry, Makelov, Schmidt, Tsipras, and Vladu]{madry2017towards}
Madry, A., Makelov, A., Schmidt, L., Tsipras, D., and Vladu, A.
\newblock Towards deep learning models resistant to adversarial attacks.
\newblock \emph{arXiv preprint arXiv:1706.06083}, 2017.

\bibitem[McAllester(2003)]{mcallester2003simplified}
McAllester, D.
\newblock Simplified pac-bayesian margin bounds.
\newblock In \emph{Learning Theory and Kernel Machines: 16th Annual Conference on Learning Theory and 7th Kernel Workshop, COLT/Kernel 2003, Washington, DC, USA, August 24-27, 2003. Proceedings}, pp.\  203--215. Springer, 2003.

\bibitem[McAllester(1999)]{mcallester1999pac}
McAllester, D.~A.
\newblock Pac-bayesian model averaging.
\newblock In \emph{Proceedings of the twelfth annual conference on Computational learning theory}, pp.\  164--170, 1999.

\bibitem[Mehra et~al.(2021)Mehra, Kailkhura, Chen, and Hamm]{mehra2021robust}
Mehra, A., Kailkhura, B., Chen, P.-Y., and Hamm, J.
\newblock How robust are randomized smoothing based defenses to data poisoning?
\newblock In \emph{CVPR}, 2021.

\bibitem[Mohapatra et~al.(2020)Mohapatra, Ko, Weng, Chen, Liu, and Daniel]{mohapatra2020higher}
Mohapatra, J., Ko, C.-Y., Weng, T.-W., Chen, P.-Y., Liu, S., and Daniel, L.
\newblock Higher-order certification for randomized smoothing.
\newblock \emph{NeurIPS}, 2020.

\bibitem[Mohapatra et~al.(2021)Mohapatra, Ko, Weng, Chen, Liu, and Daniel]{mohapatra2021hidden}
Mohapatra, J., Ko, C.-Y., Weng, L., Chen, P.-Y., Liu, S., and Daniel, L.
\newblock Hidden cost of randomized smoothing.
\newblock In \emph{International Conference on Artificial Intelligence and Statistics}, 2021.

\bibitem[Morvant et~al.(2012)Morvant, Ko{\c{c}}o, and Ralaivola]{morvant2012pac}
Morvant, E., Ko{\c{c}}o, S., and Ralaivola, L.
\newblock Pac-bayesian generalization bound on confusion matrix for multi-class classification.
\newblock In \emph{ICML}, 2012.

\bibitem[Mu et~al.(2023)Mu, Ruan, Marcolino, Jin, and Ni]{mu2023certified}
Mu, R., Ruan, W., Marcolino, L.~S., Jin, G., and Ni, Q.
\newblock Certified policy smoothing for cooperative multi-agent reinforcement learning.
\newblock In \emph{AAAI}, 2023.

\bibitem[Neyshabur et~al.(2017{\natexlab{a}})Neyshabur, Bhojanapalli, McAllester, and Srebro]{neyshabur2017exploring}
Neyshabur, B., Bhojanapalli, S., McAllester, D., and Srebro, N.
\newblock Exploring generalization in deep learning.
\newblock \emph{arXiv preprint arXiv:1706.08947}, 2017{\natexlab{a}}.

\bibitem[Neyshabur et~al.(2017{\natexlab{b}})Neyshabur, Bhojanapalli, and Srebro]{neyshabur2017pac}
Neyshabur, B., Bhojanapalli, S., and Srebro, N.
\newblock A pac-bayesian approach to spectrally-normalized margin bounds for neural networks.
\newblock \emph{arXiv preprint arXiv:1707.09564}, 2017{\natexlab{b}}.

\bibitem[Nichol \& Dhariwal(2021)Nichol and Dhariwal]{nichol2021improved}
Nichol, A.~Q. and Dhariwal, P.
\newblock Improved denoising diffusion probabilistic models.
\newblock In \emph{ICML}, 2021.

\bibitem[Papernot et~al.(2016)Papernot, McDaniel, Wu, Jha, and Swami]{papernot2016distillation}
Papernot, N., McDaniel, P., Wu, X., Jha, S., and Swami, A.
\newblock Distillation as a defense to adversarial perturbations against deep neural networks.
\newblock In \emph{2016 IEEE symposium on security and privacy (SP)}, pp.\  582--597. IEEE, 2016.

\bibitem[Rosenfeld et~al.(2020)Rosenfeld, Winston, Ravikumar, and Kolter]{rosenfeld2020certified}
Rosenfeld, E., Winston, E., Ravikumar, P., and Kolter, Z.
\newblock Certified robustness to label-flipping attacks via randomized smoothing.
\newblock In \emph{ICML}, 2020.

\bibitem[Salman et~al.(2019)Salman, Li, Razenshteyn, Zhang, Zhang, Bubeck, and Yang]{salman2019provably}
Salman, H., Li, J., Razenshteyn, I., Zhang, P., Zhang, H., Bubeck, S., and Yang, G.
\newblock Provably robust deep learning via adversarially trained smoothed classifiers.
\newblock \emph{Advances in Neural Information Processing Systems}, 32, 2019.

\bibitem[Schuchardt \& G{\"u}nnemann(2022)Schuchardt and G{\"u}nnemann]{schuchardt2022invariance}
Schuchardt, J. and G{\"u}nnemann, S.
\newblock Invariance-aware randomized smoothing certificates.
\newblock \emph{NeurIPS}, 2022.

\bibitem[Schuchardt et~al.(2020)Schuchardt, Bojchevski, Gasteiger, and G{\"u}nnemann]{schuchardt2020collective}
Schuchardt, J., Bojchevski, A., Gasteiger, J., and G{\"u}nnemann, S.
\newblock Collective robustness certificates: Exploiting interdependence in graph neural networks.
\newblock In \emph{ICLR}, 2020.

\bibitem[Sun et~al.(2023)Sun, Xu, Yao, Liang, Wu, Liang, Liu, and Liu]{sun2023improving}
Sun, C., Xu, C., Yao, C., Liang, S., Wu, Y., Liang, D., Liu, X., and Liu, A.
\newblock Improving robust fariness via balance adversarial training.
\newblock In \emph{AAAI}, 2023.

\bibitem[Tram{\`e}r et~al.(2017)Tram{\`e}r, Kurakin, Papernot, Goodfellow, Boneh, and McDaniel]{tramer2017ensemble}
Tram{\`e}r, F., Kurakin, A., Papernot, N., Goodfellow, I., Boneh, D., and McDaniel, P.
\newblock Ensemble adversarial training: Attacks and defenses.
\newblock \emph{arXiv preprint arXiv:1705.07204}, 2017.

\bibitem[Uesato et~al.(2018)Uesato, O’donoghue, Kohli, and Oord]{uesato2018adversarial}
Uesato, J., O’donoghue, B., Kohli, P., and Oord, A.
\newblock Adversarial risk and the dangers of evaluating against weak attacks.
\newblock In \emph{International Conference on Machine Learning}, pp.\  5025--5034. PMLR, 2018.

\bibitem[Wang et~al.(2021)Wang, Jia, Cao, and Gong]{wang2021certified}
Wang, B., Jia, J., Cao, X., and Gong, N.~Z.
\newblock Certified robustness of graph neural networks against adversarial structural perturbation.
\newblock In \emph{SIGKDD}, 2021.

\bibitem[Wang et~al.(2023)Wang, Pang, Du, Lin, Liu, and Yan]{wang2023better}
Wang, Z., Pang, T., Du, C., Lin, M., Liu, W., and Yan, S.
\newblock Better diffusion models further improve adversarial training.
\newblock In \emph{ICML}, 2023.

\bibitem[Weber et~al.(2023)Weber, Xu, Karla{\v{s}}, Zhang, and Li]{weber2023rab}
Weber, M., Xu, X., Karla{\v{s}}, B., Zhang, C., and Li, B.
\newblock Rab: Provable robustness against backdoor attacks.
\newblock In \emph{2023 IEEE Symposium on Security and Privacy (SP)}, 2023.

\bibitem[Wei et~al.(2023)Wei, Wang, Guo, and Wang]{wei2023cfa}
Wei, Z., Wang, Y., Guo, Y., and Wang, Y.
\newblock Cfa: Class-wise calibrated fair adversarial training.
\newblock In \emph{CVPR}, 2023.

\bibitem[Wong \& Kolter(2018)Wong and Kolter]{wong2018provable}
Wong, E. and Kolter, Z.
\newblock Provable defenses against adversarial examples via the convex outer adversarial polytope.
\newblock In \emph{ICML}, 2018.

\bibitem[Wong et~al.(2018)Wong, Schmidt, Metzen, and Kolter]{wong2018scaling}
Wong, E., Schmidt, F., Metzen, J.~H., and Kolter, J.~Z.
\newblock Scaling provable adversarial defenses.
\newblock \emph{Advances in Neural Information Processing Systems}, 31, 2018.

\bibitem[Wu et~al.(2020{\natexlab{a}})Wu, Wang, Xia, Bailey, and Ma]{wu2020skip}
Wu, D., Wang, Y., Xia, S.-T., Bailey, J., and Ma, X.
\newblock Skip connections matter: On the transferability of adversarial examples generated with resnets.
\newblock \emph{arXiv preprint arXiv:2002.05990}, 2020{\natexlab{a}}.

\bibitem[Wu et~al.(2020{\natexlab{b}})Wu, Xia, and Wang]{DBLP:conf/nips/WuX020}
Wu, D., Xia, S., and Wang, Y.
\newblock Adversarial weight perturbation helps robust generalization.
\newblock In \emph{NeurIPS}, 2020{\natexlab{b}}.

\bibitem[Xie et~al.(2021)Xie, Chen, Chen, and Li]{xie2021crfl}
Xie, C., Chen, M., Chen, P.-Y., and Li, B.
\newblock Crfl: Certifiably robust federated learning against backdoor attacks.
\newblock In \emph{International Conference on Machine Learning}, pp.\  11372--11382. PMLR, 2021.

\bibitem[Xu et~al.(2021)Xu, Liu, Li, Jain, and Tang]{xu2021robust}
Xu, H., Liu, X., Li, Y., Jain, A., and Tang, J.
\newblock To be robust or to be fair: Towards fairness in adversarial training.
\newblock In \emph{ICML}, 2021.

\bibitem[Xu et~al.(2017)Xu, Evans, and Qi]{xu2017feature}
Xu, W., Evans, D., and Qi, Y.
\newblock Feature squeezing: Detecting adversarial examples in deep neural networks.
\newblock \emph{arXiv preprint arXiv:1704.01155}, 2017.

\bibitem[Yang et~al.(2020)Yang, Duan, Hu, Salman, Razenshteyn, and Li]{yang2020randomized}
Yang, G., Duan, T., Hu, J.~E., Salman, H., Razenshteyn, I., and Li, J.
\newblock Randomized smoothing of all shapes and sizes.
\newblock In \emph{ICML}, 2020.

\bibitem[Yoshida \& Miyato(2017)Yoshida and Miyato]{yoshida2017spectral}
Yoshida, Y. and Miyato, T.
\newblock Spectral norm regularization for improving the generalizability of deep learning.
\newblock \emph{arXiv preprint arXiv:1705.10941}, 2017.

\bibitem[Zagoruyko \& Komodakis(2016)Zagoruyko and Komodakis]{DBLP:conf/bmvc/ZagoruykoK16}
Zagoruyko, S. and Komodakis, N.
\newblock Wide residual networks.
\newblock In \emph{BMVC}, 2016.

\bibitem[Zhang et~al.(2020)Zhang, Ye, Gong, Zhu, and Liu]{dinghuaizhang}
Zhang, D., Ye, M., Gong, C., Zhu, Z., and Liu, Q.
\newblock Black-box certification with randomized smoothing: A functional optimization based framework.
\newblock In \emph{NeurIPS}, 2020.

\bibitem[Zhang et~al.(2019)Zhang, Yu, Jiao, Xing, El~Ghaoui, and Jordan]{zhang2019theoretically}
Zhang, H., Yu, Y., Jiao, J., Xing, E., El~Ghaoui, L., and Jordan, M.
\newblock Theoretically principled trade-off between robustness and accuracy.
\newblock In \emph{ICML}, 2019.

\bibitem[Zhang et~al.(2024)Zhang, Zhang, Mu, Huang, and Ruan]{zhang2024towards}
Zhang, Y., Zhang, T., Mu, R., Huang, X., and Ruan, W.
\newblock Towards fairness-aware adversarial learning.
\newblock \emph{arXiv preprint arXiv:2402.17729}, 2024.

\end{thebibliography}
\bibliographystyle{icml2025}

%%%%%%%%%%%%%%%%%%%%%%%%%%%%%%%%%%%%%%%%%%%%%%%%%%%%%%%%%%%%%%%%%%%%%%%%%%%%%%%
%%%%%%%%%%%%%%%%%%%%%%%%%%%%%%%%%%%%%%%%%%%%%%%%%%%%%%%%%%%%%%%%%%%%%%%%%%%%%%%
% APPENDIX
%%%%%%%%%%%%%%%%%%%%%%%%%%%%%%%%%%%%%%%%%%%%%%%%%%%%%%%%%%%%%%%%%%%%%%%%%%%%%%%
%%%%%%%%%%%%%%%%%%%%%%%%%%%%%%%%%%%%%%%%%%%%%%%%%%%%%%%%%%%%%%%%%%%%%%%%%%%%%%%
\newpage
\appendix
\onecolumn

\section{Proof for \Cref{lem:main1}}
\label{app:pf1}

We follow the proof framework of \citet{neyshabur2017pac},
let $\mathcal{U}$ be the set of perturbations with the following property:
\begin{equation}
\label{eq:su}
\mathcal{U} \subseteq\left\{(\mathbf{u},\vv)\Big|\max _{\mathbf{x} \in \mathcal{X}_B}| f_{\mathbf{w}}(\mathbf{x+v})-\left.f_{\mathbf{w}}(\mathbf{x})\right|_{\infty}<\frac{\gamma}{8} \;\cap\; 
\max _{\mathbf{x} \in \mathcal{X}_B}| f_{\mathbf{w+u}}(\mathbf{x})-\left.f_{\mathbf{w}}(\mathbf{x})\right|_{\infty}<\frac{\gamma}{8}
\right\}.
\end{equation}
Let $q$ be the joint probability density function over $\ul, \vv$. 
We construct a new distribution $\tilde Q$ over $\tilde \ul, \tilde \vv$ that is restricted to $\mathcal{U}$ with the probability density function:
\begin{equation}
\label{eq:qu}
\tilde{q}(\tilde{\mathbf{u}},\tilde{\mathbf{v}})= \begin{cases} \frac{1}{z} q(\tilde{\mathbf{u}},\tilde{\mathbf{v}}) & \tilde{\mathbf{u}},\tilde{\mathbf{v}} \in \mathcal{U}, \\ 0 & \text { otherwise}, \end{cases}
\end{equation}
where $z$ is a normalizing constant and by the lemma assumption $z=\pr((\ul,\vv)\in \mathcal{U}\ge\frac{1}{2}$.
By the definition of $\tilde Q$, we have:
\begin{equation}
\max _{\mathbf{x} \in \mathcal{X}_B} | f_{\mathbf{w}}(\mathbf{x}+\tilde \vv)-\left.f_{\mathbf{w}}(\mathbf{x})\right|_{\infty}<\frac{\gamma}{8} \;\text{and}\;
\max _{\mathbf{x} \in \mathcal{X}_B} | f_{\mathbf{w}+\tilde \ul}(\mathbf{x})-\left.f_{\mathbf{w}}(\mathbf{x})\right|_{\infty}<\frac{\gamma}{8}.
\end{equation}
Therefore, with probability at least $1-\delta$ over training dataset $\St$, we have:
\begin{equation}\nonumber
\begin{aligned}
\|\C^{\tilde f_{\w,\vv}}_\D \|_2 &\le \|\C^{\tilde Q}_{\D,\frac{\gamma}{2}}\|_2 \quad\quad\quad\quad\quad\quad & \triangleright \text{because of Proof~\ref{lem:app1}} & \\
&\le \|\C^{\tilde Q}_{\St,\frac{\gamma}{2}}\|_2 + \sqrt{\frac{8 d_y}{m_{min} - 8d_y} \left[ \KL(\w+\tilde\ul \| P) + \ln \left( \frac{m_{min}}{4\delta} \right) \right]}   & \triangleright \text{because of Thm.~\ref{thm:2.1}} & \\
&\le \|\C^{\tilde f_{\w,\vv}}_{\St,\gamma}\|_2 + \sqrt{\frac{8 d_y}{m_{min} - 8d_y} \left[ \KL(\w+\tilde\ul \| P) + \ln \left( \frac{m_{min}}{4\delta} \right) \right]}   & \triangleright \text{because of Proof~\ref{lem:app2}} & \\
&\le \|\C^{\tilde f_{\w,\vv}}_{\St,\gamma}\|_2 + 4\sqrt{\frac{8 d_y}{m_{min} - 8d_y} \left[ \KL(\w+\ul \| P) + \ln \left( \frac{3 m_{min}}{4\delta} \right) \right]}  & \triangleright \text{because of Proof~\ref{pf:b4}} & 
\end{aligned}
\end{equation}
Hence, proved. \hfill $\square$

\begin{mypro}
\label{lem:app1}
Given (\ref{eq:su}) and (\ref{eq:qu}), for all $(\tilde \ul, \tilde\vv) \in \tilde Q$, we have
\begin{equation}
\label{eq:app1.1}
\begin{aligned}
&\max_{\x\in\X}|f_{\mathbf{w}}(\mathbf{x}+\tilde \vv)-f_{\mathbf{w}}(\mathbf{x})|_\infty<\frac{\gamma}{8},\\
&\max_{\x\in\X}|f_{\mathbf{w}+\tilde\ul}(\mathbf{x})-f_{\mathbf{w}}(\mathbf{x})|_\infty<\frac{\gamma}{8}.
\end{aligned}
\end{equation}
For all $\x\in \X$ s.t. $\tilde f_{\w,\vv}(\x)\ne y$, as $z=\pr((\ul,\vv)\in \mathcal{U})\ge\frac{1}{2}$, there exists $(\check\ul,\check\vv) \in \tilde Q$ s.t.
\begin{equation}
f_{\mathbf{w}}(\mathbf{x}+\check \vv)[\tilde f_{\w,\vv}(\x)]>f_{\mathbf{w}}(\mathbf{x}+\check \vv)[y].    
\end{equation}

Then for all $(\tilde \ul, \tilde \vv) \in \tilde Q$,  all $\x\in \X$ s.t. $\tilde f_{\w,\vv}(\x)\ne y$, we have 
\begin{equation}
\begin{aligned}
f_{\mathbf{w}+\tilde \ul}(\mathbf{x})[\tilde f_{\w,\vv}(\x)]+\frac{\gamma}{4} &> f_{\mathbf{w}}(\mathbf{x})[\tilde f_{\w,\vv}(\x)]+\frac{\gamma}{8}\\
&> f_{\mathbf{w}}(\mathbf{x}+\check \vv)[\tilde f_{\w,\vv}(\x)]\\
&>f_{\mathbf{w}}(\mathbf{x}+\check \vv)[y]\\
&>f_{\mathbf{w}}(\mathbf{x})[y]-\frac{\gamma}{8}\\
&>f_{\mathbf{w}+\tilde \ul}(\mathbf{x})[y]-\frac{\gamma}{4}.
\end{aligned}
\end{equation}
Thus for all $i\ne j$, we have
\begin{equation}
    (\C^{\tilde f_{\w,\vv}}_\D)_{ij} \le (\C^{\tilde Q}_{\D,\frac{\gamma}{2}})_{ij}.
\end{equation}
According to Perron–Frobenius theorem \citep{frobenius1912matrizen}, for all $1\le i,j\le d_y$, $\frac{\partial\|\C\|_2}{\partial (\C)_{ij}}\ge 0$.
Combine the above conditions, we get $\|\C^{\tilde f_{\w,\vv}}_\D \|_2 \le \|\C^{\tilde Q}_{\D,\frac{\gamma}{2}}\|_2$. 
\hfill $\square$
\end{mypro}

\begin{mypro}
\label{lem:app2}
Given  (\ref{eq:su}), (\ref{eq:qu}) and (\ref{eq:app1.1}), for all $(\tilde \ul, \tilde \vv) \in \tilde Q$, 
if there exists $\x\in\X$ and $(\check \ul, \check \vv) \in \tilde Q$ s.t. 
$f_{\w+\check \ul}(\x)[y]<\max_{j\ne y}f_{\w+\check \ul}(\x)[j]+\frac{\gamma}{2}$, 
we have
\begin{equation}
\begin{aligned}
f_{\mathbf{w}}(\mathbf{x}+\tilde \vv)[j]+\frac{3\gamma}{4}&>f_\w(\x)[j]+\frac{5\gamma}{8}\\
&>f_{\w+\check \ul}(\x)[j]+\frac{\gamma}{2}\\
&>f_{\w+\check \ul}(\x)[y]\\
&>f_{\w}(\x)[y]-\frac{\gamma}{8}\\
&>f_{\mathbf{w}+\tilde \ul}(\mathbf{x})[y]-\frac{\gamma}{4}.
\end{aligned}
\end{equation}
Hence, if there exists $\x\in\X$ and $(\check \ul, \check \vv) \in \tilde Q$ s.t. 
$f_{\w+\check \ul}(\x)[y]<\max_{j\ne y}f_{\w+\check \ul}(\x)[j]+\frac{\gamma}{2}$, we have 
$f_{\mathbf{w}}(\mathbf{x}+\tilde \vv)[j]+\gamma>f_{\mathbf{w}+\tilde \ul}(\mathbf{x})[y]$ for all $(\tilde \ul, \tilde \vv) \in \tilde Q$. 
Given $z\ge \frac{1}{2}$, we have $\tilde f_{\w,\vv}(\x,\gamma)\ne y$. 
Thus, $\forall i\ne j$, we have $(\C^{\tilde Q}_{\St,\frac{\gamma}{2}})_{ij}\le(\C^{\tilde f_{\w,\vv}}_{\St,\gamma})_{ij}$.
According to Perron–Frobenius theorem, we get $\|\C^{\tilde Q}_{\St,\frac{\gamma}{2}}\|_{2}\le\|\C^{\tilde f_{\w,\vv}}_{\St,\gamma}\|_{2}$.
\hfill $\square$
\end{mypro}

\begin{mypro}
\label{pf:b4}
Given $q$, $\tilde q$, $z$, and $\mathcal{U}$ in (\ref{eq:qu}),
let $\mathcal{U}^c$ denote the complement set of $\mathcal{U}$ and $\tilde q^c$ denote the normalized density function restricted to $\mathcal{U}^c$. 
Then, we have
\begin{equation}
\KL(q\| p) = z\KL(\tilde q\| p) + (1-z)\KL(\tilde q^c\| p)-H(z),
\end{equation}
where $H(z)=-z \ln z-(1-z) \ln (1-z) \leq 1$ is the binary entropy function. 
Since $\KL$ is always positive, we get 
\begin{equation}
\KL(\tilde{q} \| p)=\frac{1}{z}[\KL(q \| p)+H(z))-(1-z) \KL(\tilde{q}^c \| p)] \leq 2(\KL(q \| p)+1).
\end{equation}
Thus we have $2(\KL(\w+\ul || P)+\ln \frac{3m_{min}}{4\delta})\ge \KL(\w+\tilde\ul || P)+\ln \frac{m_{min}}{4\delta}$.
\hfill $\square$
\end{mypro}

\section{Proof for \Cref{lem:main2}}
\label{app:pf2}

The proof of Lem.~\ref{lem:main2} follows a two-step approach developed by \citet{neyshabur2017pac}. 
First, we determine the maximum permissible magnitude of perturbation $\ul$ that satisfies the margin constraint $\gamma$. 
Second, we compute the KL divergence term in the bound using this derived perturbation limit. 
This sequential analysis enables us to establish the PAC-Bayesian bound.

Consider a neural network with weights $\mathbf{W}$ that can be regularized by dividing each weight matrix $\mathbf{W}_l$ by its spectral norm $\|\mathbf{W}_l\|_2$. We define $\beta$ as the geometric mean of the spectral norms across all weight matrices:
$$\beta = \left(\prod_{l=1}^n \|\mathbf{W}_l\|_2\right)^{\frac{1}{n}},$$
where $n$ denotes the number of weight matrices. Using $\beta$, we construct normalized weights $\widetilde{\mathbf{W}}_l$ by scaling each original matrix $\mathbf{W}_l$:
$$\widetilde{\mathbf{W}}_l = \frac{\beta}{\|\mathbf{W}_l\|_2} \mathbf{W}_l.$$
Due to the homogeneity property of ReLU activation functions, the network with normalized weights $f_{\widetilde{\mathbf{w}}}$ exhibits identical behavior to the original network $f_\mathbf{w}$.

A key observation is that the product of spectral norms remains invariant under our normalization, with $\prod_{l=1}^n \|\mathbf{W}_l\|_2 = \prod_{l=1}^n \|\widetilde{\mathbf{W}}_l\|_2$. Additionally, the ratio of Frobenius to spectral norms is preserved for each layer:
$$\frac{\|\mathbf{W}_l\|_F}{\|\mathbf{W}_l\|_2} = \frac{\|\widetilde{\mathbf{W}}_l\|_F}{\|\widetilde{\mathbf{W}}_l\|_2}.$$
Since this normalization preserves the excess error stated in the theorem, we can proceed with the proof using only the normalized weights $\widetilde{\mathbf{w}}$. Thus, without loss of generality, we assume $\|\mathbf{W}_l\|_2 = \beta$ for all layers $l$.

In our approach, we define the prior distribution $P$ as a Gaussian distribution with zero mean and diagonal covariance matrix $\sigma^2 \mathbf{I}$. We introduce random perturbations $\mathbf{u} \sim \mathcal{N}(0, \sigma^2 \mathbf{I})$, where $\sigma$ is determined later in relation to $\beta$. To maintain prior independence from the learned predictor $\mathbf{w}$ and its norm, we select $\sigma$ based on an estimated value $\tilde{\beta}$.
The PAC-Bayesian bound is computed for each $\tilde{\beta}$ chosen from a pre-determined grid, providing generalization guarantees for all $\mathbf{w}$ that satisfy $|\beta - \tilde{\beta}| \leq \frac{1}{n} \beta$. This grid construction ensures coverage of all relevant $\beta$ values by some $\tilde{\beta}$. We then apply a union bound across all grid-defined $\tilde{\beta}$ values.
For our analysis, we consider the set of $\tilde{\beta}$ and corresponding $\mathbf{w}$ satisfying $|\beta - \tilde{\beta}| \leq \frac{1}{n} \beta$, which implies:
$$\frac{1}{e} \beta^{n-1} \leq \tilde{\beta}^{n-1} \leq e \beta^{n-1}.$$

Leveraging results from \citet{bandeira2021spectral} and the Gaussian perturbation of $\mathbf{u} \sim \mathcal{N}(0, \sigma^2 \mathbf{I})$, we derive a probabilistic bound for the spectral norm of the perturbation matrix $\mathbf{U}_l$ ($\ul_l = \vecc(\mathbf{U}_l)$):
\begin{equation}
\label{eq:uwbound}
\mathbb{P}_{\mathbf{u}_l \sim \mathcal{N}(0, \sigma^2 \mathbf{I})}\left[\|\mathbf{U}_l\|_2 > t\right] \leq 2h \exp\left(-\frac{t^2}{2h\sigma^2}\right),
\end{equation}
where $h$ denotes the hidden layer width. Applying a union bound across all layers, we establish that the spectral norm of perturbation $\mathbf{U}_l$ in each layer is bounded by $\sigma \sqrt{2h \ln (4 nh)}$ with probability at least $\frac{1}{2}$.

Combining the bound with Lem.~\ref{lem:peturbound}, we get
\begin{equation}
\label{eq:cod1}
\begin{aligned}
\max _{\mathbf{x} \in \mathcal{X}_B}\left\|f_{\mathbf{w}+\mathbf{u}}(\mathbf{x})-f_{\mathbf{w}}(\mathbf{x})\right\|_2 & \leq e B \beta^n \sum_l \frac{\left\|\mathbf{U}_l\right\|_2}{\beta} \\
& =e B \beta^{n-1} \sum_l\left\|\mathbf{U}_l\right\|_2 \\
&\leq e^2 n B \tilde{\beta}^{n-1} \sigma \sqrt{2 h \ln (4 n h)} \leq \frac{\gamma}{8}.
\end{aligned}
\end{equation}

To satisfy (\ref{eq:cod1}), consider $\tilde{\beta}^{n-1} \leq e \beta^{n-1}$, the largest $\sigma$ can be chosen as 
\begin{equation}\nonumber
\sigma = \frac{\gamma}{228 n B \sqrt{h \ln (4 n h)}\prod_{l=1}^n \|\W_l\|_2^{\frac{n-1}{n}}}.
\end{equation}

Using this choice of $\sigma$, the perturbation $\mathbf{u}$ satisfies the conditions required by Lemma \ref{lem:main1}. Next, we evaluate the KL divergence term by computing it explicitly for our chosen prior distribution $P$ and posterior distribution $Q$,
\begin{equation}\nonumber
\begin{aligned}
\KL(\w+\ul \| P) &\le \frac{\|\w\|_2^2}{2\sigma^2}\\
&=\frac{\sum_{l=1}^n\|\W_l\|_F^2}{2\sigma^2}\\
&\le \mathcal{O}\left(B^2n^2h\ln(nh)\frac{\prod_{l=1}^n \|\W_l\|_2^2}{\gamma^2} \sum_{l=1}^n \frac{\|\W_l\|^2_F}{\|\W_l\|^2_2} \right).
\end{aligned}
\end{equation}
Finally, we establish a union bound across different values of $\tilde \beta$. Following \citet{neyshabur2017pac}, we need only consider the range $\left(\frac{\gamma}{2 B}\right)^{\frac{1}{n}} \leq \beta \leq\left(\frac{\gamma \sqrt{m}}{2 B}\right)^{\frac{1}{n}}$, which can be covered by a grid of size $nm^{\frac{1}{2n}}$. Consequently, with probability at least $1-\delta$, the following bound holds for any $\tilde \beta$ and all $\w$ satisfying $|\beta-\tilde{\beta}| \leq \frac{1}{n} \beta$:
\begin{equation}
\|\C^{\tilde f_{\w,\vv}}_\D \|_2 \leq \| \C^{\tilde f_{\w,\vv}}_{\St,\gamma}\|_2 + \mathcal{O}\left(\sqrt{\frac{d_y}{(m_{min} - 8d_y)\gamma^2} \left[ \Phi(f_\w) + \ln \left( \frac{n m_{min}}{\delta} \right) \right]}\right),
\end{equation}
where $\Phi(f_\w)=B^2n^2h\ln(nh)\prod_{l=1}^n ||\W_l||_2^2 \sum_{l=1}^n \frac{||\W_l||^2_F}{||\W_l||^2_2}$.

Hence, proved. \hfill $\square$

\begin{lemma}[\citet{neyshabur2017pac}]
\label{lem:peturbound}
For any $B, n > 0$, let $f_{\mathbf{w}}: \mathcal{X}_B$ $\rightarrow \mathcal{Y}$ be a $n$-layer feedforward network with ReLU activation function.
Then for any $\w$, and $\x\in \X$, and any perturbation $\ul=\vecc(\{\mathbf{U}_l\}_{l=1}^n)$ such that $\|\mathbf{U}_l\|_2\le \frac{1}{n}\|\W_l\|_2$, the change in the output of the network can be bounded as follow
\begin{equation}
\begin{aligned}
\left\|f_{\mathbf{w}+\mathbf{u}}(\mathbf{x})-f_{\mathbf{w}}(\mathbf{x})\right\|_2 \leq e B\left(\prod_{l=1}^n\left\|\W_l\right\|_2\right) \sum_{l=1}^n \frac{\|\mathbf{U}_l\|_2}{\left\|\W_l\right\|_2}.
\end{aligned}
\end{equation}
\end{lemma}

\section{Randomized Smoothing Algorithm}
Randomized smoothing \citep{cohen2019certified} was developed to evaluate probabilistic certified robustness for classification tasks. 
It aims to construct a smoothed model $\tilde f_{\w,\vv}(\x)$, which can produce the most probable prediction of the base classifier $f_\w(\x)$ over perturbed inputs from Gaussian noise in a test instance. 
The smoothed classifier $\tilde f_{\w,\vv}(\x)$ is supposed to be provably robust to $\ell_2$-norm bounded perturbations within a certain radius.
%\begin{tcolorbox}[colback=blue(back), colframe=blue(back)]
\begin{theorem}\label{the:cohen}
 \cite{cohen2019certified} For a classifier $f: \mathbb{R} \to \mathcal{Y}$, suppose $c \in \mathcal{Y}$, let $\vv \sim \mathcal{N}(0, \sigma_{v}^2 I)$, the smoothed classifier be 
 $\tilde f_{\w,\vv}(\x):=\mathop{\arg\max}\limits_c \mathbb{P}(f(\x+\vv)=c)$,
 suppose $\underline{p_a},\overline{p_b} \in [0,1]$, if
 \begin{equation}
     \mathbb{P} (f(\x+\vv)=c_a)\geq \underline{p_a} \geq \overline{p_b} \geq \mathop{\max}\limits_{c\neq c_a}\mathbb{P}(f(\x+\vv)=c),
 \end{equation}
 then $\tilde f_{\w,\vv}(\x+\epsilon)=c_a$ for all $\|\epsilon \|_2 \leq R$, where
 \begin{equation}
     R=\frac{\sigma_v}{2}(\Psi^{-1}(\underline{p_a})-\Psi^{-1}(\overline{p_b})).
 \end{equation}
\end{theorem}
Here $\Psi^{-1}$ is the inverse cumulative distribution function (CDF) of the normal distribution.

To apply the randomized smoothing algorithm in practice, the following steps are involved.

\subsection{Experimental Setup and Sampling}
Base Classifier: A base classifier $f_\w(\x)$, such as a deep neural network, is trained on the task of interest. This model will serve as the foundation for smoothing.

Noise Sampling: For a given test input $\x$, perturbations $\vv \sim \mathcal{N}(0, \sigma_v^2 I)$ are generated by sampling from a Gaussian distribution with mean 0 and variance $\sigma_v^2$. These perturbations are added to the input to simulate noise in the environment.

Prediction Averaging: The model $f_\w(\x)$ is evaluated on a set of perturbed inputs, $\x + \vv_i$ where $\vv_i \sim \mathcal{N}(0, \sigma^2 I)$ for $i = 1, 2, \dots, N$. The smoothed classifier $\tilde f_{\w,\vv}(\x)$ is determined by computing the majority vote (or the most probable class) across the set of predictions for each perturbed input:

$$\tilde f_{\w,\vv}(\x) = \mathop{\arg\max}_c \sum_{i=1}^N \mathbbm{1}[\mathop{\arg\max}_j f(\x + \vv_i)[j] = c], $$
where $\mathbbm{1}$ is the indicator function.

Certifiable Robustness: The goal is to determine how robust the smoothed model $\tilde f_{\w,\vv}(\x)$ is to $\ell_2$-norm perturbations. 
The perturbation radius $R$ is computed using the theorem above, which depends on the probabilities of the classifier $f_\w$ on the noisy inputs and the inverse CDF of the normal distribution. This gives a certifiable radius of robustness within which the classifier $\tilde f_{\w,\vv}(\x)$ is guaranteed to output the correct class.

Performance Evaluation: The effectiveness of randomized smoothing can be evaluated by testing the classifier's accuracy under various perturbation levels. Typically, experiments involve evaluating the smoothed model $\tilde f_{\w,\vv}(\x)$ against a range of $\epsilon$ values (perturbation magnitudes) to assess the robustness of the classifier across different noise levels.

\end{document}